%% file: root.tex
\documentclass[letterpaper, 10 pt, journal, twoside]{IEEEtran}
\ifCLASSINFOpdf
\else
\fi
\input{doc/preamble}

\begin{document}
\title{Online Extrinsic Calibration based on Per-Sensor Ego-Motion Using Dual Quaternions}

% \author{Michael~Shell,~\IEEEmembership{Member,~IEEE,}
%         John~Doe,~\IEEEmembership{Fellow,~OSA,}
%         and~Jane~Doe,~\IEEEmembership{Life~Fellow,~IEEE}% <-this % stops a space
% \thanks{M. Shell was with the Department
% of Electrical and Computer Engineering, Georgia Institute of Technology, Atlanta,
% GA, 30332 USA e-mail: (see http://www.michaelshell.org/contact.html).}% <-this % stops a space
% \thanks{J. Doe and J. Doe are with Anonymous University.}% <-this % stops a space
% \thanks{Manuscript received April 19, 2005; revised August 26, 2015.}}
\author{Markus Horn$^{*}$, Thomas Wodtko$^{*}$, Michael Buchholz and Klaus Dietmayer%
\thanks{Manuscript received: October 10, 2020; Revised January 8, 2021; Accepted January 18, 2021.}%
\thanks{This paper was recommended for publication by Editor Lucia Pallottino upon evaluation of the Associate Editor and Reviewers' comments.} %
\thanks{All authors are with the Institute of Measurement, Control and Microtechnology, Ulm University, Albert-Einstein-Allee 41, 89081 Ulm, Germany {\tt\footnotesize \{firstname\}.\{lastname\}@uni-ulm.de}}%
%\thanks{Digital Object Identifier (DOI): see top of this page.}%
\thanks{$^{*}$ \textit{Markus Horn and Thomas Wodtko are co-first authors. Corresponding author: Markus Horn.}}
}

\markboth{IEEE Robotics and Automation Letters. Preprint Version. Accepted January, 2021}
{Horn, Wodtko \MakeLowercase{\textit{et al.}}: Online Extrinsic Calibration based on Per-Sensor Ego-Motion Using Dual Quaternions} 




\maketitle

\begin{abstract}
In this work, we propose an approach for extrinsic sensor calibration from per-sensor ego-motion estimates.
Our problem formulation is based on dual quaternions, enabling two different online capable solving approaches.
We provide a certifiable globally optimal and a fast local approach along with a method to verify the globality of the local approach.
Additionally, means for integrating previous knowledge, for example, a common ground plane for planar sensor motion, are described.
Our algorithms are evaluated on simulated data and on a publicly available dataset containing RGB-D camera images.
Further, our online calibration approach is tested on the KITTI odometry dataset, which provides data of a lidar and two stereo camera systems mounted on a vehicle.
Our evaluation confirms the short run time, state-of-the-art accuracy, as well as online capability of our approach while retaining the global optimality of the solution at any time.
\end{abstract}

% \begin{IEEEkeywords}
% IEEE, IEEEtran, journal, \LaTeX, paper, template.
% \end{IEEEkeywords}

% \fi
\IEEEpeerreviewmaketitle

% content
\input{doc/1_introduction}
\input{doc/2_related_work}

\input{doc/3_problem_formulation}
\input{doc/4_algorithms}
\input{doc/5_experiments}
\input{doc/6_conclusion}
\input{doc/7_appendix}

\ifCLASSOPTIONcaptionsoff
  \newpage
\fi



%\bibliographystyle{IEEEtran}
%\bibliography{mybibfile}
%\begin{thebibliography}{1}


%\end{thebibliography}
\printbibliography


%\end{IEEEbiography}


%\newpage



%\vfill


\end{document}

%% file: doc/preamble.tex
\usepackage{algorithm}
\usepackage{algorithmicx}  % algorithmic environment
\usepackage[noend]{algpseudocode}
\usepackage{amsfonts}
\usepackage{amsmath}  % commands for dealing with mathematics
\usepackage{bm}
\usepackage{booktabs}
\usepackage{etoolbox}
\usepackage{graphicx}  % required if you want graphics, photos, etc.
\usepackage{hyperref}  % hyperlinks
\usepackage{mathtools}
\usepackage{multirow}
\usepackage{siunitx}
\usepackage{stfloats}  % double column floats
\usepackage{tabu}

\usepackage{tikz}
\usetikzlibrary{calc}
\usetikzlibrary{arrows,backgrounds,fit,positioning}
\usetikzlibrary{plotmarks}

\usepackage{tikzscale}
\usepackage{threeparttablex}
\usepackage{url}  % better support for handling and breaking URLs

\usepackage{pgfplots}   
\pgfplotsset{compat=newest}
\usepgfplotslibrary{patchplots} % draws high-quality function plots in normal or logarithmic scaling
\usepgfplotslibrary{groupplots}
\usepackage{grffile}

\definecolor{githubColor}{HTML}{2EA44F}
\newcommand{\gitref}[2]{\href{#1}{\color{githubColor}{#2}}}%

\definecolor{newGray}{HTML}{808080}
\newcommand{\makegray}[1]{\color{newGray}#1}%

\newcolumntype{O}[1]{S[detect-weight, mode=text, table-format=#1]}

\usepackage[
backend=biber,
bibencoding=utf8,
style=ieee,
giveninits=true,
doi=false,
isbn=false,
url=false,
sorting=none
]{biblatex}
\AtBeginBibliography{\footnotesize}

\graphicspath{{img/}}

\hypersetup{colorlinks=false}

\renewcommand{\bfseries}{\fontseries{b}\selectfont} % <--
\robustify\bfseries             % <--
\newrobustcmd{\B}{\bfseries} 

\newcommand{\norm}[1]{\left\lVert#1\right\rVert}
\newcommand{\vmnorm}[1]{\ensuremath \left\| #1 \right\|}

\newcommand{\mbeq}{\overset{!}{=}}

\renewcommand{\vec}[1]{\boldsymbol{#1}}

\newcommand{\sol}[1]{\hat{#1}}

\newcommand{\mat}[1]{\mathbf{#1}}
\newcommand{\trans}{^\text{\rmfamily \textup{T}}}
\newcommand{\argmin}{\operatorname{arg\,min}}
\newcommand{\diag}{\operatorname{diag}}
\newcommand{\nullspace}{\operatorname{null}}
\newcommand{\vecspan}{\operatorname{span}}
\newcommand{\vectorize}{\operatorname{vec}}

\newcommand{\pspace}{\,}

\newcommand\copyrighttext{\footnotesize \textcopyright~2021 IEEE. Personal use of this material is permitted. Permission from IEEE must be obtained for all other uses, in any current or future media, including reprinting/republishing this material for advertising or promotional purposes, creating new collective works, for resale or redistribution to servers or lists, or reuse of any copyrighted component of this work in other works.
DOI: \href{https://ieeexplore.ieee.org/document/9345480}{10.1109/LRA.2021.3056352}
}%

\newcommand\copyrightnotice{%
	\begin{tikzpicture}[remember picture,overlay]
	\node[anchor=south] at (current page.south) {\fbox{\parbox{\dimexpr\textwidth-\fboxsep-\fboxrule\relax}{\copyrighttext}}};
	\end{tikzpicture}%
}

%% file: doc/1_introduction.tex
% !TEX root = ../root.tex

\section{Introduction}
% for IEEE journal papers produced under \LaTeX\ using
% IEEEtran.cls version 1.8b and later.

\copyrightnotice
\IEEEPARstart{W}{ith} an evolving automation process, a growing number of sensors are embedded in robotic systems.
Two exemplary reasons for this are increasing the field of view (FoV) or densifying certain areas.
While intrinsic sensor parameters are often determined independently for each sensor, an extrinsic calibration is the key for fusing data from various sensors.
Thus, extrinsic sensor calibration is a crucial step in the process of building any robotic system.

Several approaches for extrinsic sensor calibration have been proposed so far.
While following the same objective, they distinctively vary in strategy.
Most available algorithms belong to one of the following three categories: target-based, registration-based, and motion-based.
Target-based algorithms \cite{geiger2012automatic, kummerle2018automatic, domhof2019extrinsic} aim to detect engineered calibration targets within the FoV of multiple sensors at different positions or points in time.
The extrinsic calibration is then estimated by matching the estimated target poses within each sensor's coordinate system.
In contrast, registration-based approaches \cite{besl1992method, levinson2013automatic, schneider2017regnet} directly use overlapping FoVs of multiple sensors and estimate the calibration by aligning their measurements.
For example, for lidar-camera calibration \cite{levinson2013automatic}, lidar depth discontinuities are aligned with image edges.
However, target- and registration-based methods are subject to severe limitations, since they either require a specific calibration procedure, involving capturing the target at different positions, or can only be used on sensors with overlapping FoVs.
Furthermore, they are restricted to specific sensor types and cannot be used on motion sensors like Inertial Measurement Units (IMUs).
In contrast, motion-based calibration approaches \cite{brookshire2013extrinsic, giamou2019certifiably}, including our proposed algorithm, only require per-sensor ego-motion estimates.
This makes them suitable for many sensor types such as cameras, lidars, radars, or IMUs.
Although per-sensor ego-motion is not directly available from most sensors, algorithms used for registration and mapping are effective tools to estimate it \cite{zhang2014loam, sumikura2019openvslam, horn2020deepclr}.
Since we want to offer a universal calibration tool that can be deployed on many different sensor setups, our work is focused on motion-based extrinsic calibration.

\begin{figure}[t]
    \centering
    \def\svgwidth{0.95\columnwidth}
    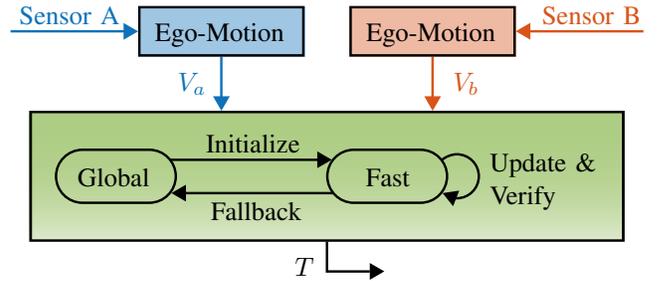
    \caption{Our proposed extrinsic calibration can be divided into distinct parts.
    First, ego-motions $V_a$ and $V_b$ are estimated separately for each sensor.
    The resulting motions are then passed to our calibration, which consists of a fast local and a global estimation part for estimating the calibration $T$.
    In the beginning, a global solution is calculated.
    This global solution is then used to initialize the fast optimization.
    When updating the fast solution, its globality is verified by evaluating the duality gap.
    In case of errors, the global solution is used as a fallback.}
    \label{fig:blockDiagramm}
\end{figure}

While some calibration approaches are designed for online use \cite{levinson2013automatic,schneider2013odometry}, most algorithms are developed with regard to offline calibration.
However, using online calibration, new sensors can be easily integrated into the system without extensive calibration procedures.
Further, online calibration can be used to verify and correct a given calibration.
This is an important aspect since sensors of robotic systems can be displaced, for example, after maintenance.
Fig.~\ref{fig:blockDiagramm} gives an overview of our online calibration approach. 

One of the most diverse aspects when dealing with calibration is the representation of transformations and poses in 3D space.
They have 6 degrees of freedom~(DOF)~\cite{mccarthy1990introduction}, though different representations are often either highly over-represented or cumbersome to calculate with.
Mostly, homogeneous matrices or dual quaternions~(DQs) are used~\cite{mccarthy1990introduction}, both over-representing 6 DOF. 
While Euler angles are a more compact representation for orientations, without further definitions, their representation has many ambiguities.
Additionally, they are not directly feasible for error calculation since, for example, $L^2$ errors can differ when changing the defined rotation axes order.
Therefore, Euler angle representations should be avoided whenever possible.
Homogeneous matrices for rigid body displacements are another way for representing 3D transformations.
Although they are better for calculation, as they form a group w.r.t. multiplications, they are still subject to constraints and ambiguities.
Those need close attention when working with them.
Especially for optimization approaches, these constraints must always be enforced, leading to cumbersome problem definitions.

In contrast, DQs provide a rather compact representation and still form a group w.r.t. multiplications.
Furthermore, only a single ambiguity concerning the sign exists for DQs, and only two constraints are necessary to ensure a valid transformation representation.
Hence, we formulate our problem based on DQs, since they are more feasible and compact than homogeneous matrices and over-represent rigid displacements by only two parameters.

Our main contributions are:
\begin{itemize}
    \item a problem formulation allowing integration of previous knowledge in Section~\ref{sec:problemFormulation}, e.g., common ground planes for planar motion,
    \item a globally optimal online extrinsic calibration approach from per-sensor ego-motion, based on a DQ optimization problem in Section~\ref{sec:algorithms},
    \item an extensive evaluation based on simulated and on real-world data from public datasets in Section~\ref{sec:experiments},
    \item a Python-based open-source library for ego-motion simulation and extrinsic calibration\footnote{\gitref{https://github.com/uulm-mrm/motion_based_calibration}{https://github.com/uulm-mrm/motion\_based\_calibration}}.
\end{itemize}

%\IEEEpubidadjcol

%% file: img/overview.pdf_tex
%% Accompanies image file 'overview_new.pdf' (pdf, eps, ps)
\begingroup%
  \makeatletter%
  \providecommand\color[2][]{%
    \errmessage{(Inkscape) Color is used for the text in Inkscape, but the package 'color.sty' is not loaded}%
    \renewcommand\color[2][]{}%
  }%
  \providecommand\transparent[1]{%
    \errmessage{(Inkscape) Transparency is used (non-zero) for the text in Inkscape, but the package 'transparent.sty' is not loaded}%
    \renewcommand\transparent[1]{}%
  }%
  \providecommand\rotatebox[2]{#2}%
  \newcommand*\fsize{\dimexpr\f@size pt\relax}%
  \newcommand*\lineheight[1]{\fontsize{\fsize}{#1\fsize}\selectfont}%
  \ifx\svgwidth\undefined%
    \setlength{\unitlength}{196.84275055bp}%
    \ifx\svgscale\undefined%
      \relax%
    \else%
      \setlength{\unitlength}{\unitlength * \real{\svgscale}}%
    \fi%
  \else%
    \setlength{\unitlength}{\svgwidth}%
  \fi%
  \global\let\svgwidth\undefined%
  \global\let\svgscale\undefined%
  \makeatother%
  \begin{picture}(1,0.43693253)%
    \lineheight{1}%
    \setlength\tabcolsep{0pt}%
    \put(0,0){\includegraphics[width=\unitlength,page=1]{overview.pdf}}%
    \put(0.10615225,0.14874934){\makebox(0,0)[lt]{\lineheight{1.25}\smash{\begin{tabular}[t]{l}Global\end{tabular}}}}%
    \put(0.5618181,0.14874934){\makebox(0,0)[lt]{\lineheight{1.25}\smash{\begin{tabular}[t]{l}Fast\end{tabular}}}}%
    \put(0,0){\includegraphics[width=\unitlength,page=2]{overview.pdf}}%
    \put(0.31660881,0.09540727){\makebox(0,0)[lt]{\lineheight{1.25}\smash{\begin{tabular}[t]{l}Fallback\end{tabular}}}}%
    \put(0.30873742,0.2014551){\makebox(0,0)[lt]{\lineheight{1.25}\smash{\begin{tabular}[t]{l}Initialize\end{tabular}}}}%
    \put(0,0){\includegraphics[width=\unitlength,page=3]{overview.pdf}}%
    \put(0.22824425,0.37799069){\makebox(0,0)[lt]{\lineheight{1.25}\smash{\begin{tabular}[t]{l}Ego-Motion\end{tabular}}}}%
    \put(0.56202654,0.37799069){\makebox(0,0)[lt]{\lineheight{1.25}\smash{\begin{tabular}[t]{l}Ego-Motion\end{tabular}}}}%
    \put(0.26462659,0.29893012){\color[rgb]{0.05490196,0.44705882,0.7254902}\makebox(0,0)[lt]{\lineheight{1.25}\smash{\begin{tabular}[t]{l}$V_a$\end{tabular}}}}%
    \put(0.69929626,0.29893012){\color[rgb]{0.84705882,0.32941176,0.09803922}\makebox(0,0)[lt]{\lineheight{1.25}\smash{\begin{tabular}[t]{l}$V_b$\end{tabular}}}}%
    \put(0,0){\includegraphics[width=\unitlength,page=4]{overview.pdf}}%
    \put(0.44931855,0.00586878){\makebox(0,0)[lt]{\lineheight{1.25}\smash{\begin{tabular}[t]{l}$T$\end{tabular}}}}%
    \put(0.83986215,0.40466172){\color[rgb]{0.84705882,0.32941176,0.09803922}\makebox(0,0)[lt]{\lineheight{1.25}\smash{\begin{tabular}[t]{l}Sensor B\end{tabular}}}}%
    \put(0.01400964,0.40466172){\color[rgb]{0.05490196,0.44705882,0.7254902}\makebox(0,0)[lt]{\lineheight{1.25}\smash{\begin{tabular}[t]{l}Sensor A\end{tabular}}}}%
    \put(0,0){\includegraphics[width=\unitlength,page=5]{overview.pdf}}%
    \put(0.75755504,0.17247716){\color[rgb]{0,0,0}\makebox(0,0)[lt]{\lineheight{1.25}\smash{\begin{tabular}[t]{l}Update \&\\Verify\end{tabular}}}}%
  \end{picture}%
\endgroup%

%% file: doc/2_related_work.tex
% !TEX root = ../root.tex

\section{Related Work}

For estimating the required per-sensor ego-motion, registration algorithms \cite{besl1992method, horn2020deepclr} or online capable SLAM algorithms like LOAM for lidars~\cite{zhang2014loam} or OpenVSLAM for cameras~\cite{sumikura2019openvslam} can be used.
Additionally, SLAM algorithms are often directly used to estimate the sensor calibration by registering the individual maps of multiple per-sensor SLAM algorithms~\cite{carrera2011slam, heng2013camodocal, wang2017online}.
Nevertheless, overlaps between the maps are crucial in this case.
This creates unwanted requirements on the ego-motion, for example, large rotations within a small position range~\cite{carrera2011slam}.

However, many other algorithms only rely on per-sensor ego-motion.
Most of them use the rotation matrix or the quaternion representation for the rotation and handle the translation separately as a vector.
This also means, similar to~\cite{taylor2015motion} or~\cite{holtz2017automatic}, the rotation is often optimized first, and afterward, the translation in a consecutive step.
In~\cite{holtz2017automatic}, the quaternion representation for rotations is used, and quaternion products are transformed into matrix-vector products for defining the optimization problem.
Eigenvalue decomposition is then applied for obtaining the solution.

In contrast, DQs provide a combined representation for rotation and translation.
DQs are more widely used in hand-eye calibration tasks~\cite{daniilidis1999hand}, although approaches also exist for regular extrinsic calibration~\cite{brookshire2013extrinsic}.
In~\cite{daniilidis1999hand}, the problem is solved analytically by applying the Singular Value Decomposition~(SVD).
Assuming the two smallest singular values are almost zero, the associated right-singular vectors are then projected onto the constrained manifold.
However, as our experiments indicate, this approach fails to find an appropriate solution in case this assumptions is violated, i.e., for planar motion with higher noise.
In~\cite{brookshire2013extrinsic}, the optimization problem is defined based on the Lie algebra of the Lie group of DQs and the calibration is solved by a non-linear least squares optimization.

A major drawback of the previously mentioned works is their local optimization approach, which leads to dependencies of the result on the initialization.
Addressing this issue, based on~\cite{briales2017convex}, a global optimization for extrinsic and hand-eye calibration with additional scale estimation is proposed by~\cite{giamou2019certifiably, wise2020certifiably}.
The rotation matrix and the translation vector are estimated using Lagrange duality.
However, due to the rotation matrix representation, the optimization has 22 constraints in total, which increases the computation time.
Similarly, another global approach proposed in~\cite{huang2017extrinsic} suffers from a high computational burden, more precisely, \SI{6}{\second} for only 1630 transformation pairs.
In contrast to the previously mentioned approaches, we provide online capable global and local strategies based on the DQ representation and an additional globality verification for the local approach.

For the special case of indoor robotics and autonomous vehicles with planar motion, not all calibration parameters are observable since the rotation axes of all transformations are parallel~\cite{brookshire2013extrinsic}.
Therefore, the non-observable parameter $z$ as well as the roll and pitch angle are often estimated from a common ground plane~\cite{heng2013camodocal, holtz2017automatic, zuniga2019automatic} and only the remaining parameters $x$, $y$, and yaw angle are estimated from per-sensor motion.
Since our approach is based on DQs, we can restrict the optimization on planar calibration only, i.e., $x$, $y$, and yaw angle, for the cost of only two additional constraints.

To summarize this section, to the best of our knowledge, we propose the first online calibration with certifiably global solutions, which can be used for non-planar and planar motion-based calibration.

%% file: doc/3_problem_formulation.tex
% !TEX root = ../root.tex

\section{Problem Formulation}
\label{sec:problemFormulation}

In this section, we derive the optimization problem similar to~\cite{briales2017convex, giamou2019certifiably}, but represent rigid displacements with DQs instead of homogeneous matrices.
This results in a simplified problem formulation involving fewer constraints.
Thus, DQs are generally introduced before the actual objective function is derived.

For our work, we assume synchronized sensor pairs and known intrinsic calibrations.
This also includes the scaling for monocular camera motion estimation.
If sensors do not provide simultaneous measurements but provide their data in a common time frame, the motion estimations can be interpolated, for example, similar to~\cite{zuniga2019automatic}.
Further, all noise, if not stated differently, is assumed to originate from additive white Gaussian processes.

In the following, transformations are generally referred to as functions, denoted with uppercase letters, meaning no representation is implicitly specified.
A composition of two transformations $T_a$ and $T_b$ is denoted by $T_{ab} = T_a \circ T_b$, which refers to the usual function composition.
(Dual) quaternions are denoted with lowercase letters and the dot multiplication $``\cdot"$ refers to the respective (dual) quaternion multiplication.
Vectorized representations and matrices are always denoted with lowercase and uppercase bold letters, respectively. 
For example, if $q$ is a (dual) quaternion, $\vec{q} = \vectorize(q)$ represents $q$ as a vector.
Thus, the transformation $T_{ab} = T_a \circ T_b$ can be described using homogeneous matrices $\mat{T}_{ab} = \mat{T_b} \mat{T_a}$ or DQs $q_{ab} = q_b \cdot q_a$.

\subsection{Dual Quaternions}
\label{subsec:dualQuat}

DQs, which combine the dual number theory with ordinary quaternions, are briefly introduced in this section.
For a more detailed explanation, we refer to~\cite{mccarthy1990introduction}.
Given two quaternions $q_r$ and $q_d$, a DQ $q$ is noted by
\begin{equation}
    q = q_r + \epsilon \, q_d \pspace ,
\end{equation}
where $\epsilon$ is the dual number with $\epsilon^2 = 0$ and $\epsilon \ne 0$, which commutes with all other elements of the algebra.
$q_r$ is called the real part and $q_d$ the dual part. 
The conjugate of a DQ $q^*$ is defined as
\begin{equation}
    q^* = q_r^* + \epsilon \, q_d^* \pspace ,
\end{equation}
with $q_r^*$ and $q_d^*$ being the respective conjugated quaternions.
Similar to quaternions, a DQ $q$ is a unit DQ if and only if
\begin{equation}
\label{unitDualQuat}
    \norm{q}^2 = q \cdot q^* \mbeq 1 \pspace .
\end{equation}
Carrying out the multiplication of (\ref{unitDualQuat}) leads to
\begin{align}
    \norm{q}^2 = q \cdot q^* & = (q_r + \epsilon \, q_d) \cdot (q_r^* + \epsilon \, q_d^*) \nonumber \\
    & =  \norm{q_r}^2 + \epsilon (q_r \cdot q_d^* + q_d \cdot q_r^*) \mbeq 1 \pspace .
\end{align}
Thus, a DQ $q = q_r + \epsilon \, q_d$ is unit if and only if
\begin{equation}
\label{eq:unitDualQuat}
    \norm{q_r}^2 = 1 \quad \land \quad q_r \cdot q_d^* + q_d \cdot q_r^* = 0 \pspace .
\end{equation}

The group of unit DQs is of special interest for this work, since it surjectively represents rigid body displacements, as explained in the following.
By restricting the real part of DQs to be positive, rigid body displacements are represented unambiguously in our work.
Given an arbitrary rotation with axis $\vec{n}\in\mathbb{R}^3$ and magnitude $\theta\in\mathbb{R}$, as well as a translation $\vec{t}\in\mathbb{R}^3$, the quaternions $r$ and $t$, representing rotation and translation, are created first by 
\begin{equation}
\label{eq:rt_quaternions}
r = \left(\text{cos}\frac{\theta}{2},\,\vec{n} \, \text{sin}\frac{\theta}{2}\right) \pspace , \quad t = (0,\vec{t}) \pspace ,
\end{equation}
using the scalar-vector representation of quaternions.
The DQ $q$ is subsequently defined by
\begin{equation}
    \label{eq:dq_from_tr}
    q = r + \epsilon \cdot \frac{1}{2}\cdot t \cdot r \pspace ,
\end{equation}
representing the respective 3D transformation of $r$ and $t$.
Such a transformation can be applied to a point $\vec{v}\in\mathbb{R}^3$ similar to ordinary quaternions.
To do so, the respective DQ $q_v = 1 + \epsilon \, (0,\vec{v})$ is transformed by
\begin{equation}
    \Tilde{q_v} = q \cdot q_v \cdot q^* \pspace .
\end{equation}
The transformed point $\Tilde{\vec{v}}$ can be retrieved from $\Tilde{q_v}$ since the structure is preserved.

A key feature of DQs, which we will take advantage of later, is that multiplications can be expressed using matrices.
Given two DQs $p$ and $q$, their multiplication is defined by
\begin{equation}
\label{eq:dualQuatMultiplication}
    p \cdot q = p_r \cdot q_r + \epsilon \, (p_r \cdot q_d + p_d \cdot q_r) \pspace .
\end{equation}
Let $\mat{M}^l_{p_r},\mat{M}^l_{p_d}$ and $\mat{M}^r_{q_r},\mat{M}^r_{q_d}$ represent the left and right side multiplications of the respective ordinary quaternions~\cite{mccarthy1990introduction}. Now, $\mat{Q}^l_p$ and $\mat{Q}^r_q$ defined by
\begin{equation}
    \mat{Q}^l_p = \begin{bmatrix}
        \mat{M}^l_{p_r} & \mat{0}\\
        \mat{M}^l_{p_d} & \mat{M}^l_{p_r}
    \end{bmatrix} \pspace , \quad
    \mat{Q}^r_q = \begin{bmatrix}
        \mat{M}^r_{q_r} & \mat{0}\\
        \mat{M}^r_{q_d} & \mat{M}^r_{q_r}
    \end{bmatrix} \pspace ,
\end{equation}
represent the left and right side multiplication of $p$ and $q$, respectively.
Given $\mat{Q}^l_p$ and $\mat{Q}^r_q$, the multiplication \eqref{eq:dualQuatMultiplication} can be expressed with
\begin{equation}
    \vectorize(p \cdot q) = \mat{Q}^l_p \cdot \vectorize(q) = \mat{Q}^r_q \cdot \vectorize(p) \pspace ,
\end{equation}
where $\vectorize(\cdot)$ denotes the representation of a DQ as eight dimensional vector.
Generally, quaternions are sequentialized by the order $(w,x,y,z)$, where $w$ is the real part and $(x,y,z)$ the respective vector part.
For DQs, the real and the dual vectors are concatenated.

\subsection{Deriving a Cost Function}

\begin{figure}
    \centering
    \def\svgwidth{0.95\columnwidth}
    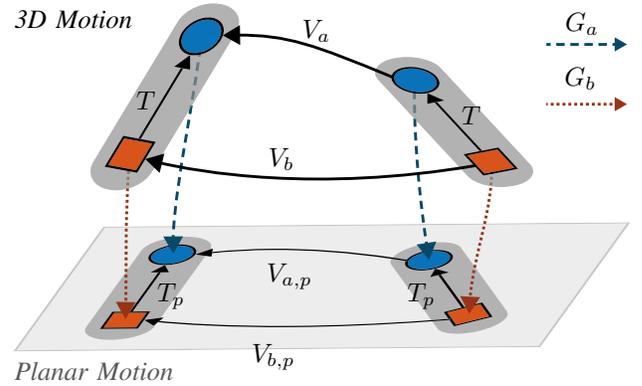
    \caption{
    Motions $V_a$ and $V_b$ of two rigidly mounted sensors with calibration $T$:
    The upper part describes general motion in 3D space, while the lower part represents the respective projection onto a plane using $G_a$ and $G_b$.}
    \label{fig:sensorMotion}
\end{figure}

Similarly to~\cite{giamou2019certifiably}, our first objective is to formulate a quadratic cost function.
Given an arbitrary graph of transformations, the loop-closing condition requires the composition of all transformations along any path with identical start and end point to result in an identity transformation.
Fig.~\ref{fig:sensorMotion} shows two rigidly connected sensors, which are freely moved in 3D space.
This motion of two sensors can be conceived as a graph, where $T$ is the transformation from sensor $B$ to sensor $A$, and $V_a$ and $V_b$ are respective sensor ego-motion transformations.
The general loop-closing condition is then given by
\begin{equation}
\label{eq:loopClosingCondition}
    V_b \circ T \circ V_a^{-1} \circ T^{-1} = I \pspace ,
\end{equation}
where $I$ denotes the identity transformation.
Based on this condition, the authors of \cite{giamou2019certifiably} formulate a quadratic function using homogeneous transformation matrices. 
In this work, we use DQs instead. 
Thus, given DQs $q_T$, $q_a$, and $q_b$ representing the respective transformations, the condition \eqref{eq:loopClosingCondition} can be written as
\begin{equation}
\label{eq:loopClosingDualQuat}
    q_a \cdot q_T = q_T \cdot q_b \pspace .
\end{equation}
Subsequently, a quadratic representation is derived by
\begin{align}
\label{eq:quadForm}
               & q_T \cdot q_b - q_a \cdot q_T &= 0 & \nonumber \\
    \iff \quad & (\mat{Q}_{q_b}^r - \mat{Q}_{q_a}^l) \, \vectorize(q_T) &= \vec{0} & \nonumber \\
    \stackrel{\norm{\cdot}_2^2}{\iff} \quad & \vectorize(q_T)\trans \, \mat{Q} \, \vectorize(q_T) &= 0 & \pspace ,
\end{align}
where the cost matrix $\mat{Q} = (\mat{Q}_{q_b}^r - \mat{Q}_{q_a}^l)\trans(\mat{Q}_{q_b}^r - \mat{Q}_{q_a}^l)$ is used.
In order to apply weights, a weight matrix $\mat{W} = \diag (\sqrt{w_1},...,\sqrt{w_8})$ can be introduced, leading to a weighted $\mat{Q}_w$ given by
\begin{equation}
    \label{eq:dataMatrix}
    \mat{Q}_w = (\mat{Q}_{q_b}^r - \mat{Q}_{q_a}^l)\trans\mat{W}\trans\mat{W}(\mat{Q}_{q_b}^r - \mat{Q}_{q_a}^l) \pspace .
\end{equation}
The previous equations are only derived for a single pair of ego-motion estimates.
For multiple steps $n$, the weighted cost matrix of step $i$ is defined as $\mat{Q}_{w,i}$. 
Let now the cost function $J:D\rightarrow\mathbb{R}$ be of the form
\begin{equation}
    \label{eq:costFunction}
    J(\vec{q}) = \!\sum_{i=1}^{n}\eta_i\,\vec{q}\trans\mat{Q}_{w,i}\vec{q} \pspace ,
\end{equation}
with weights $\eta_i$ and $\vec{q}\in\mathbb{R}^8$ representing a vectorized DQ.
$J$ represents the sum of the left side of \eqref{eq:quadForm} for $n$ steps and, thus, can be used as quadratic cost function for the following DQ based optimization procedures.
The step specific weights $\eta_i$ must satisfy $\sum_{i=1}^{n} \eta_i = 1$.
This normalization is required for a consistent globality verification, as described later in Section~\ref{sec:globalVerification}.
For $\eta_i = \frac{1}{n}$, weights are equally distributed.

\subsection{Unit Dual Quaternion Constraints}

To ensure that the DQ $q$ represents a rigid body displacement, certain constraints must be satisfied. 
Given the condition \eqref{eq:unitDualQuat}, two equality constraint functions $g_1$ and $g_2$, formally given by
\begin{subequations}
\label{eq:unitDQConstraints}
\begin{alignat}{2}
    g_1(\vec{q}) & = 1 - \norm{\vec{q}_{\{1 \dots 4\}}}^2            \mbeq 0 \pspace, \\
    g_2(\vec{q}) & = 2 (q_1q_5 + q_2q_6 + q_3q_7 + q_4q_8)  \mbeq 0 \pspace ,
\end{alignat}
\end{subequations}
are defined. 
For $\vec{q} = \begin{bmatrix} q_1 & \dots & q_8 \end{bmatrix}\trans$, the subvector containing the first four elements is denoted by $\vec{q}_{\{1 \dots 4\}}$.
The constraint $g_1$ requires the first half of $\vec{q}$ to be normalized, which reflects the left part of \eqref{eq:unitDualQuat}.
The right part is represented by $g_2$ after carrying out the multiplication.

\subsection{Optimization Problem}

Given the cost function $J$ and the two equality constraints $g_1$ and $g_2$, the optimization problem is defined by
\begin{subequations}
\label{eq:optiProb}
\begin{alignat}{2}
    & \!\min_{\vec{q}} & ~ & J(\vec{q}) \\
    & \text{w.r.t.}	 & ~ & \vec{g}(\vec{q}) =
        \begin{pmatrix} 
        1 - \norm{\vec{q}_{\{1 \dots 4\}}}^2 \\
        2(q_1q_5 + q_2q_6 + q_3q_7 + q_4q_8)
        \end{pmatrix} \mbeq \vec{0} \pspace .
\end{alignat}
\end{subequations}
In comparison to the primal optimization problem of~\cite{giamou2019certifiably}, the number of equality constraints is reduced from 22 down to 2.
Additionally, its dimension is reduced from 10 to 8.
Thus, starting from the same initial condition, the optimization problem \eqref{eq:optiProb} is a more compact representation due to the use of DQs.

A main feature of~\cite{giamou2019certifiably} is the certifiably global solution. 
In fact, the globality of a solution can only be guaranteed under certain conditions, which are outlined later. 
Otherwise, it is only possible to verify the globality of a given solution.
Since the structure of both optimization problems using either homogeneous matrices or DQs is equal, the Lagrangian dual problem can be defined similarly to~\cite{briales2017convex} for our problem \eqref{eq:optiProb}.
Furthermore, due to the small number of constraints, the respective Lagrangian function is less complicated.
Formally, given the Langrange multiplier $\vec{\lambda} = \begin{bmatrix} \lambda_1 & \lambda_2 \end{bmatrix}\trans$, it is given by
\begin{equation}
    L(\vec{q},\vec{\lambda}) = \vec{q}\trans\mat{Q}\vec{q} + \vec{\lambda}\trans\vec{g}(\vec{q}) \pspace .
\end{equation}
Using $\mat{P}_{||}(\lambda)$ and $\mat{P}_\times(\lambda)$ defined by
\begin{equation}
    \label{eq:lagrange_constraints}
    \mat{P}_{||}(\lambda) = 
    \begin{bmatrix}
        - \lambda\textbf{I}_4 & \vec{0}_4\\
        \vec{0}_4 & \vec{0}_4
    \end{bmatrix} \pspace , \quad
    \mat{P}_\times(\lambda) = 
    \begin{bmatrix}
        \vec{0}_4 & \lambda\textbf{I}_4 \\
        \lambda\textbf{I}_4  &  \vec{0}_4
    \end{bmatrix} \pspace ,
\end{equation}
it can be further simplified to
\begin{align}
    L(\vec{q},\vec{\lambda}) & = \vec{q}\trans\mat{Q}\vec{q} + \vec{q}\trans\mat{P}_{||}(\lambda_1)\vec{q} + \lambda_1 + \vec{q}\trans\mat{P}_{\times}(\lambda_2)\vec{q}, \nonumber \\
    & = \vec{q}\trans\mat{Z}(\vec{\lambda})\vec{q} + \lambda_1  \pspace ,
\end{align}
where $\mat{Z}(\vec{\lambda})$ is given by $\mat{Z}(\vec{\lambda}) = \mat{Q} + \mat{P}_{||}(\lambda_1) + \mat{P}_{\times}(\lambda_2)$.
Next, the Lagragian dual problem is specified.
It is given by
\begin{subequations}
\label{eq:dualOptiProb}
\begin{alignat}{2}
&\!\max_{\lambda_1}  &\quad& \lambda_1\\
&\text{w.r.t.} &      & \mat{Z}(\vec{\lambda}) \succeq \mat{0}  \pspace .
\end{alignat}
\end{subequations}
The dual problem results in a Semidefinite Programming (SDP) problem, for which solvers like~\cite{grant2020cvx} are available.

\subsection{Planar Motion Constraints}

In this section, we demonstrate how additional knowledge can be integrated into our proposed approaches by adding respective constraints.
A planar-only motion hereby serves as an example.
We assume that the common ground plane is already estimated for both sensors using concepts similar to~\cite{heng2013camodocal, holtz2017automatic, zuniga2019automatic}.
Without loss of generality, we treat the ground plane as $xy$-plane.
If a sensor's ground plane is given in Hesse normal form with normal vector $\vec{v}_G$ and distance $d_G$, we can calculate an aligning DQ $q_G$ by
\begin{subequations}
\begin{alignat}{3}
    \vec{n}_G &= \vec{v}_G \times \vec{e}_z \pspace ,\\
    \theta_G &= \arccos(\vec{v}_G\trans\,\vec{e}_z) \pspace ,\\
    \vec{t}_G &= d_G\,\vec{e}_z \pspace ,
\end{alignat}
\end{subequations}
with $\vec{e}_z = \begin{bmatrix} 0 & 0 & 1\end{bmatrix}\trans$ and by using \eqref{eq:rt_quaternions} and \eqref{eq:dq_from_tr}.
In general, the transformation $q_{G}$ for sensors $a$ and $b$ is denoted by $G_{a}$ and $G_{b}$, respectively.
As illustrated in Fig~\ref{fig:sensorMotion}, the planar components of estimated motion and calibration are calculated as
\begin{subequations}
\label{eq:planarMotion}
\begin{alignat}{3}
    V_{a,p} &= G_{a}^{-1} \circ V_{a} \circ G_{a} \pspace , \\
    V_{b,p} &= G_{b}^{-1} \circ V_{b} \circ G_{b} \pspace , \\
    T_{p} &= G_{b}^{-1} \circ T \circ G_{a} \pspace .
\end{alignat}
\end{subequations}
In the following, we use $V_{a,p}$ and $V_{b,p}$ as new inputs for estimating the planar calibration $T_{p}$.
Fig.~\ref{fig:sensorMotion} illustrates the transformation from 3D motion to planar motion.

After applying the previously described transformations, the values of $z$ as well as of the roll and pitch angle are assumed to be zero.
This means that $r$ and $t$ of \eqref{eq:dq_from_tr} are given by
\begin{subequations}
\label{eq:planarQuatConstraints}
\begin{alignat}{2}
    r & \mbeq \left( \cos \frac{\theta}{2} , \, \begin{bmatrix} 0 & 0 & \sin \frac{\theta}{2} \end{bmatrix}\trans \right) \pspace , \\
    t & \mbeq \left( 0 , \, \begin{bmatrix} x & y & 0 \end{bmatrix}\trans \right) \pspace .
\end{alignat}
\end{subequations}
From \eqref{eq:dq_from_tr}, we can see that $r = q_r$ and $t = 2 q_d q_r^*$.
In combination with \eqref{eq:planarQuatConstraints}, this results in the following constraints for planar-only calibration:
\begin{equation}
    q_2^2 + q_3^2 = 0 \pspace , \quad q_1 q_8 - q_4 q_5 = 0 \pspace .
\end{equation}
These constraints are included into the Lagrangian function using $\mat{P}_{r}(\lambda)$ and $\mat{P}_{t}(\lambda)$, which are defined similarly to~\eqref{eq:lagrange_constraints}.

After estimating the planar calibration $T_{p}$, the full calibration $T$ is finally calculated as
\begin{equation}
    \label{eq:planarTo3dCalib}
    T = G_{b} \circ T_{p} \circ G_{a}^{-1} \pspace .
\end{equation}

\subsection{Globality Verification}
\label{sec:globalVerification}

Solving our problem using a local optimization approach generally cannot guarantee to result in a global optimum. 
However, in our case, it is possible to verify the globality of a local solution using the verification proposed in~\cite{carlone2015lagrangian}.

Given a local solution $\sol{\vec{q}}$, the dual problem's first-order optimality condition must be satisfied for $\sol{\vec{q}}$ in order to be a global solution. 
This is formally given by
\begin{equation}
\label{eq:optimalityCondition}
    \frac{\partial L}{\partial\vec{q}} \bigg\rvert_{\substack{
    \vec{q} \, = \, \sol{\vec{q}} \\
    \vec{\lambda} \, = \, \sol{\vec{\lambda}}
    }} = \mat{Z}(\sol{\vec{\lambda}}) \, \sol{\vec{q}} \mbeq \vec{0} \pspace .
\end{equation}
Since \eqref{eq:optimalityCondition} is an over-determined linear system of equations, it can be solved using least squares algorithms w.r.t. $\sol{\vec{\lambda}}$, given the local solution $\sol{\vec{q}}$.
If its solution is such that $\mat{Z}(\sol{\vec{\lambda}}) \, \sol{\vec{q}} = \vec{0}$ and $\mat{Z}(\sol{\vec{\lambda}}) \succeq \mat{0}$, the solution $\sol{\vec{q}}$ is globally optimal. 
Otherwise, the duality gap is evaluated and serves as a measure for the distance between local and global solutions.
Since the sum of the weights $\eta_i$ in the cost function \eqref{eq:costFunction} is normalized to one, it is possible to set a threshold for the duality gap independent of the number of measurements $n$.

%% file: img/drawing.pdf_tex
%% Creator: Inkscape 1.0.1 (3bc2e813f5, 2020-09-07), www.inkscape.org
%% Accompanies image file 'drawing.pdf' (pdf, eps, ps)
\begingroup%
  \makeatletter%
  \providecommand\color[2][]{%
    \errmessage{(Inkscape) Color is used for the text in Inkscape, but the package 'color.sty' is not loaded}%
    \renewcommand\color[2][]{}%
  }%
  \providecommand\transparent[1]{%
    \errmessage{(Inkscape) Transparency is used (non-zero) for the text in Inkscape, but the package 'transparent.sty' is not loaded}%
    \renewcommand\transparent[1]{}%
  }%
  \providecommand\rotatebox[2]{#2}%
  \newcommand*\fsize{\dimexpr\f@size pt\relax}%
  \newcommand*\lineheight[1]{\fontsize{\fsize}{#1\fsize}\selectfont}%
  \ifx\svgwidth\undefined%
    \setlength{\unitlength}{166.3125bp}%
    \ifx\svgscale\undefined%
      \relax%
    \else%
      \setlength{\unitlength}{\unitlength * \real{\svgscale}}%
    \fi%
  \else%
    \setlength{\unitlength}{\svgwidth}%
  \fi%
  \global\let\svgwidth\undefined%
  \global\let\svgscale\undefined%
  \makeatother%
  \begin{picture}(1,0.62250733)%
    \lineheight{1}%
    \setlength\tabcolsep{0pt}%
    \put(0,0){\includegraphics[width=\unitlength,page=1]{drawing.pdf}}%
    \put(0,0){\includegraphics[width=\unitlength,page=3]{drawing.pdf}}%
    \put(0.63514677,0.13530507){\makebox(0,0)[lt]{\lineheight{1.25}\smash{\begin{tabular}[t]{l}$T_p$\end{tabular}}}}%
    \put(0.41267103,0.16687215){\makebox(0,0)[lt]{\lineheight{1.25}\smash{\begin{tabular}[t]{l}$V_{a,p}$\end{tabular}}}}%
    \put(0.38903315,0.04037475){\makebox(0,0)[lt]{\lineheight{1.25}\smash{\begin{tabular}[t]{l}$V_{b,p}$\end{tabular}}}}%
    \put(0.23963742,0.13530507){\makebox(0,0)[lt]{\lineheight{1.25}\smash{\begin{tabular}[t]{l}$T_p$\end{tabular}}}}%
    \put(0.41533079,0.3498097){\makebox(0,0)[lt]{\lineheight{1.25}\smash{\begin{tabular}[t]{l}$V_b$\end{tabular}}}}%
    \put(0.46893483,0.55274092){\makebox(0,0)[lt]{\lineheight{1.25}\smash{\begin{tabular}[t]{l}$V_a$\end{tabular}}}}%
    \put(0.72089921,0.41502232){\makebox(0,0)[lt]{\lineheight{1.25}\smash{\begin{tabular}[t]{l}$T$\end{tabular}}}}%
    \put(0.20697407,0.43907644){\makebox(0,0)[lt]{\lineheight{1.25}\smash{\begin{tabular}[t]{l}$T$\end{tabular}}}}%
    \put(0,0){\includegraphics[width=\unitlength,page=2]{drawing.pdf}}%
    \put(0.88460069,0.5636451){\makebox(0,0)[lt]{\lineheight{1.25}\smash{\begin{tabular}[t]{l}$G_a$\end{tabular}}}}%
    \put(0.88360541,0.47636844){\makebox(0,0)[lt]{\lineheight{1.25}\smash{\begin{tabular}[t]{l}$G_b$\end{tabular}}}}%
    \put(0.01419391,0.01156441){\color[rgb]{0.34117647,0.34117647,0.3372549}\makebox(0,0)[lt]{\lineheight{1.25}\smash{\begin{tabular}[t]{l}\textit{\textit{Planar Motion}}\end{tabular}}}}%
    \put(0.01419391,0.56849786){\makebox(0,0)[lt]{\lineheight{1.25}\smash{\begin{tabular}[t]{l}\textit{\textit{3D Motion}}\end{tabular}}}}%
  \end{picture}%
\endgroup%

%% file: doc/4_algorithms.tex
% !TEX root = ../root.tex

\section{Algorithms}
\label{sec:algorithms}

In this section, we describe the fast local and the global optimization approach for the previously described problem formulation.
Further, we propose an algorithm which combines both approaches for globally optimal online calibration.

\subsection{Fast Optimization}
\label{sec:fast_optimization}

Since the primal optimization problem \eqref{eq:optiProb} only has two constraints, a fast optimization can be achieved by applying standard constrained quadratic solving approaches.
We have compared the performance of two approaches, i.e., Sequential Quadratic Programming (SQP) and Interior-Point (IP)~\cite{nocedal2006numerical}, in preliminary investigations.
Since SQP methods have performed better, they are used in the following.

Due to the quadratic structure of the cost function, its gradient $\nabla J(\vec{q}) = 2\mat{Q}\vec{q}$ can be passed to the optimizer analytically.
Although the fast optimization only retains local solutions, their globality can still be verified using the concept described in Section~\ref{sec:globalVerification}.
Performance differences, especially during online calibration, are investigated later in Section~\ref{sec:experiments}.

\subsection{Global Optimization}
\label{sec:global_optimization}

While local optimization approaches are well suited for certain scenarios, a global solution is mostly favored when calibration is only performed once and with lower speed requirements.
Thus, similar to~\cite{briales2017convex}, a global approach solving the Lagrangian dual problem is outlined here.
As stated before, the optimization problem given in \eqref{eq:dualOptiProb} can be solved using solvers like~\cite{grant2020cvx}.
However, these solvers generally only deliver a global dual solution $\sol{\vec{\lambda}}$, based on which the respective primal solution has to be recovered.
In general, the primal solution is given by
\begin{equation}
    \sol{\vec{q}} = \underset{\vec{q}}{\argmin} \, L(\vec{q}, \sol{\vec{\lambda}}) \pspace ,
\end{equation}
which results in a quadratic subproblem given by
\begin{equation}
    \label{eq:globalSubproblem}
    \sol{\vec{q}} = \underset{\vec{q}}{\argmin} \, \vec{q}\trans \mat{Z}(\sol{\vec{\lambda}})\vec{q} \pspace .
\end{equation}
Only if the duality gap is zero, the primal solution is guaranteed to be global.
This is achieved if and only if the subproblem can be solved such that $\sol{\vec{q}}{}\trans \mat{Z}(\sol{\vec{\lambda}})\sol{\vec{q}} = 0$ holds.
Since $\mat{Z}\succeq\mat{0}$ is required, $\sol{\vec{q}}$ has to be an element of the null space of $\mat{Z}(\sol{\vec{\lambda}})$.
Assuming a one-dimensional null space similar to~\cite{brookshire2013extrinsic}, we can use $g_1$ to retrieve a unique solution:
\begin{align}
\label{eq:dualToPrimal}
\vecspan(\{\vec{v}\}) & = \nullspace(\textbf{Z}(\sol{\vec{\lambda}})) \pspace , \nonumber \\
\sol{\vec{q}} & = \frac{\vec{v}}{\vmnorm{\vec{v}_{\{1,...,4\}}}} \pspace .
\end{align}
Since \eqref{eq:dualToPrimal} represents all possible solutions and at least one solution must exist, other constraints, e.g. $g_2$, are always satisfied. 
As stated above, since the duality gap is zero, a solution recovered this way is guaranteed to be global.

\subsection{Online Calibration}
\label{sec:onlineCalib}

For online calibration, the previously described fast and global optimization approaches are combined.
As shown in Fig.~\ref{fig:blockDiagramm} and illustrated in Algorithm~\ref{alg:onlineCalib} for planar motion, both approaches rely on the same input data.
In each iteration, the data matrix $\mat{Q}_i$ is updated and used to calculate the normalized $\mat{Q}$.
If the fast optimization yields verified global solutions for at least $t_{noFail}$, the global optimization is not invoked.
Nevertheless, the globality of the fast optimization solution is verified in every iteration and the global optimization is used as a fallback in case of discrepancies.
By this, our verification ensures global solutions, even when only the local optimization is used.
Additionally, due to the initialization with the solution of the previous iteration, our local approach converges faster, improving the execution time.
$t_{lastLocalError}$ is initialized to the start time $t_1$ since the first global solution is used to initialize the fast optimization.

\newpage
\clearpage
\vspace*{-5mm}

\begin{algorithm}
\caption{Online Extrinsic Calibration for Planar Motion}
\label{alg:onlineCalib}

\hspace*{\algorithmicindent} \textbf{Input:} Sensor motion $\mathcal{Q}_i = \{q_{a,i}, q_{b,i}\}$, weights $W_i$ \\
\hspace*{\algorithmicindent} \textbf{Constant:} Min. time w/o local error $t_{noFail}$ \\
\hspace*{\algorithmicindent} \textbf{Output:} Calibration $\sol{\vec{q}}_{i}$

\begin{algorithmic}[1]
\Procedure{Calibrate}{$\mathcal{Q}_i$, $W_i$, $t_i$, $t_{lastLocalError}$}
    \State Form $\mathcal{Q}_{i,p}$ from $\mathcal{Q}_i$  \Comment{planar motion \eqref{eq:planarMotion}}
    
    \State Form $\mat{Q}_{w,i}$ from $\mathcal{Q}_{i,p}$ and $W_i$ 
    \Comment{data matrix \eqref{eq:dataMatrix}}
    
    \State $\mat{Q}_i \gets \mat{Q}_{i-1} + \mat{Q}_{w,i}$ 
    \Comment{\eqref{eq:costFunction}}
    
    \State $\mat{Q} \gets \mat{Q}_i / i$ 
    \Comment{equally distributed $\eta_i$}
    
    \State $\sol{\vec{q}}_{i,p} \gets$ Solution with SQP (init: $\sol{\vec{q}}_{i-1,p})$  \Comment{\ref{sec:fast_optimization}}
        
    \State $isglobal \gets$ Check globality  of $\tilde{\vec{q}}$\Comment{\ref{sec:globalVerification}}
    
    \If {$\lnot isglobal$} 
        \State $t_{lastLocalError} \gets t_i$ \Comment{store error time}
    \EndIf
    
    \If { $t_i - t_{lastLocalError} \leq t_{noFail}$ }
        \State $\sol{\vec{q}}_{i,p} \gets$ Solution with SDP  \Comment{global \eqref{eq:dualOptiProb}, \eqref{eq:globalSubproblem}}
    \EndIf
    
    \State Form $\sol{\vec{q}}_{i}$ from $\sol{\vec{q}}_{i,p}$  \Comment{3D calibration \eqref{eq:planarTo3dCalib}}
    \State \Return $\sol{\vec{q}}_{i}$, $t_{lastLocalError}$
\EndProcedure
\end{algorithmic}
\end{algorithm}

%% file: doc/5_experiments.tex
% !TEX root = ../root.tex

\section{Experiments}
\label{sec:experiments}

In this section, we describe our experiments and discuss the results.
First, motion data with artificial noise was simulated for investigating the influences of noise and different dataset sizes.
Second, our work is compared to the approaches of \cite{daniilidis1999hand}, \cite{brookshire2013extrinsic} and \cite{giamou2019certifiably} on the publicly available dataset provided by the authors of~\cite{brookshire2013extrinsic}.
Finally, the KITTI odometry dataset~\cite{geiger2012autonomous} is used to verify the influence of additional planar motion constraints and to evaluate the performance of our proposed online calibration approach.

The rotation and translation errors $\varepsilon_r$ and $\varepsilon_t$ between predicted calibration $\hat{q}$ and ground-truth calibration $q_T$ are calculated as
\begin{subequations}
\begin{alignat}{2}
    \varepsilon_r &= 2 \arccos(\vec{q}_{\varepsilon,1}) ,\\
    \varepsilon_t &= \norm{2 \cdot q_{\varepsilon,d} \cdot q^*_{\varepsilon,r}} ,
\end{alignat}
\end{subequations}
with $q_\varepsilon = q_T^{-1} \cdot \hat{q}$ and $\vec{q}_\varepsilon = \vectorize(q_\varepsilon)$.
In contrast to other error metrics, these metrics describe physical quantities, i.e., rotation magnitude and translation offset.
All evaluations were run on a computer containing an ADM\,Ryzen\textsuperscript{TM}\,7\,3700X CPU and 64GB of DDR4 RAM. 
If not stated differently, optimization times are averaged over ten runs.

\subsection{Simulated Data}

During our work, we developed a framework for simulating sensor motion data.
This framework can generate transformation matrices as well as DQs.
Using this framework, extensive test data was generated, which is publicly available with our code.
In contrast to real data, a simulation allows the configuration of parameters such as noise levels.

\begin{figure}
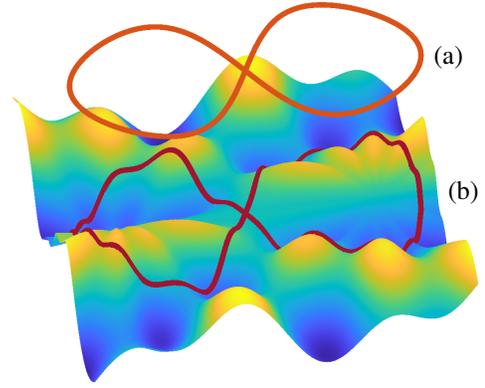

    \centering
    \include{img/surface}
    \vspace*{-5mm}
    \caption{An example of a simulated path:
    A 2D path (a) is projected onto a rough surface, resulting in a 3D path (b).
    The exemplary surface shown is a sinusoid mixture.}
    \label{fig:simulationPath}
\end{figure}

For simulation, a 2D path is projected onto a 3D surface as sketched in Fig.~\ref{fig:simulationPath}.
The surface's normal vector in combination with the path's directions define the orientations along the path. 
The poses are then used to generate transformations for rigidly connected sensors following the 6D path.

Further, unbiased Gaussian noise can be added to the transformations.
Thereby, the standard deviation is referred to as noise level. 
It can either be configured with absolute values or relative ones, generated from the average sensor motion.
For example, given a path with an average of \SI{1}{\metre} and \SI{0.1}{\radian} per transformation, a noise level of \SI{10}{\percent} leads to a respective Gaussian noise with $\sigma_{\text{trans}} = \SI{0.1}{\metre},\, \sigma_{\text{rot}} = \SI{0.01}{\radian}$.

\subsection{General Optimization Behavior}

\begin{figure}[b]
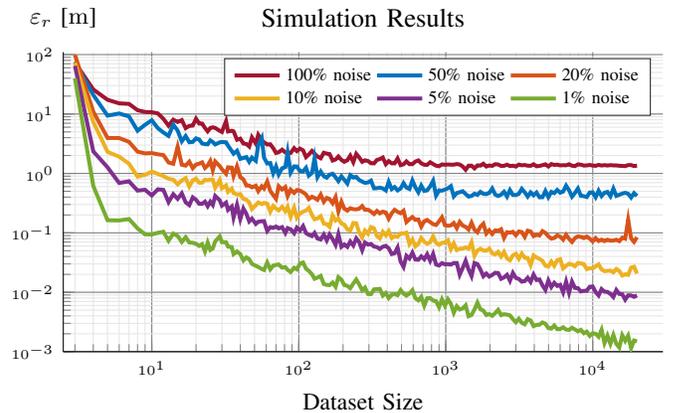

	\centering
	\include{img/translation_calib_errors}
	\caption{Translation errors of calibrations based on simulated data are plotted over the number of points used for the global optimization.
	For each data point, the median over eight simulation runs was used.
	All results are grouped by their relative noise level. Respective rotation errors tend to follow the same general course and are therefore not depicted separately.}
	\label{fig:simulatedResults}
\end{figure}

To generally test our approach, calibrations were estimated based on simulated data with different parameters. 
More precisely, the data was simulated for six noise levels, 167 dataset sizes, and eight sensor poses with distances between \SI{2}{\metre} and \SI{8}{\metre}.
Each parameter set was used eight times in total, leading to a total amount of 224\,448 sensor pair optimizations. 
The results for the global optimization are presented in Fig~\ref{fig:simulatedResults}.

Generally, the results show that rotation and translation errors decrease for either an increasing amount of points or a decreasing noise level.
Thus, we were able to verify that our approach is able to compensate for noise with an increasing amount of points.

\subsection{RGB-D Data}
\label{sec:rgbdEval}

We have evaluated and compared our approach using the data provided by the authors of~\cite{brookshire2013extrinsic}.
They provide 200 interpolated frames of per-sensor ego-motion estimates for two synchronized Xtion RGB-D cameras with full 6D-motion and known calibration.
As shown in Table~\ref{tab:rgbdData}, our approaches reach a similar translation accuracy while being considerably faster than~\cite{brookshire2013extrinsic,giamou2019certifiably}.
Although the local approach of~\cite{brookshire2013extrinsic} provides a better rotation accuracy, it takes more than 100 times longer than our local approach.
Since all assumptions of~\cite{daniilidis1999hand} are met in this experiment, this approach outperforms all other methods.
Furthermore, this experiment verifies that all methods are able to find appropriate solutions for well-conditioned data.

\begin{table}
    \centering
    \caption{Results on the RGB-D dataset.}
    \label{tab:rgbdData}
    \begin{tabular}{lO{1.2}O{1.2}O{4.1}}
         \toprule
          \textbf{Approach} & \pmb{$\varepsilon_t$} [$\si{\centi\metre}$] & \pmb{$\varepsilon_r$} [$\si{\degree}$] & \textbf{Time} [$\si{\milli\second}$]\\
         \midrule
         Analytic~\cite{daniilidis1999hand} & \B 1.08 & \B 0.94 & \B 0.2 \\
         Local~\cite{brookshire2013extrinsic} & 1.20 & 1.04 & 1900.0 \\
         Matrix~\cite{giamou2019certifiably} & 1.28 & 1.06 & 61.5 \\
         Global (ours) & 1.16 & 1.06 & 43.9 \\
         Fast (ours) & 1.16 & 1.06 & 6.2 \\
         \bottomrule
    \end{tabular}
\end{table}

\subsection{KITTI Odometry}
\label{sec:kittiOdometryEval}

\begin{figure}
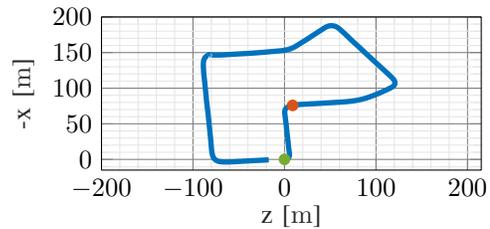
 
    \centering
    \include{img/kitti_path_07}
    \caption{The first 1000 ground-truth poses of sequence 07 of KITTI are shown.
    The first (green) and 150th (red) frame are marked with dots.}
    \label{fig:kittiSequences} 
\end{figure}

The KITTI odometry dataset \cite{geiger2012autonomous} provides calibrated data of a Velodyne HDL-64E lidar and two stereo camera setups mounted on a vehicle.
Further, ground truth poses provided by an OXTS system are available for 11 sequences.
An example is shown in Fig.~\ref{fig:kittiSequences}.
We evaluated our method on sequences 04-10, since all were recorded with the same calibration.
For this purpose, we concatenated the sequences, which resulted in a total of 12\,074 frames, covering approximately 20 minutes.
The per-sensor ego-motions of lidar and stereo cameras were estimated using online capable implementations of LOAM\footnote{ \gitref{https://github.com/HKUST-Aerial-Robotics/A-LOAM}{https://github.com/HKUST-Aerial-Robotics/A-LOAM}} and OpenVSLAM\footnote{\gitref{https://github.com/xdspacelab/openvslam}{https://github.com/xdspacelab/openvslam}}, respectively.
The ground plane for the planar motion extension was estimated in the first frame using a RANSAC based plane fitting on point clouds.

The calibration results between the lidar and the first stereo camera system are displayed in Table~\ref{tab:kittiPlaneResults}, showing that a successful extrinsic sensor calibration of autonomous vehicle sensors based on per-sensor ego-motion can be achieved.
Especially the rotation can be estimated precisely.
For the general 3D case, our approaches deliver similar results to~\cite{giamou2019certifiably} while being considerably faster.

However, the analytic solution of~\cite{daniilidis1999hand} does not yield an appropriate solution.
In contrast to the previous scenario~\ref{sec:rgbdEval}, the motion in this experiment almost only consists of planar motion, which violates the assumption of at least two non-parallel rotation axes in~\cite{daniilidis1999hand}. 
Furthermore, the sensor motion estimations are subject to relatively high noise.

Despite the use of a rather simple and unfiltered plane estimation, the results show that our planar extension leads to notable improvements.
The translation error is decreased while still providing accurate rotation estimates and, at the same time, the execution time is further reduced.

\begin{table}
\centering
\setlength{\tabcolsep}{0.37em}
\caption{Results on the KITTI odometry dataset.}
\label{tab:kittiPlaneResults}
\begin{tabular}{lO{2.2}O{1.3}O{2.1}O{2.2}O{1.3}O{2.1}}
    \toprule
    \textbf{Approach} & \multicolumn{3}{c}{\textbf{3D}} & \multicolumn{3}{c}{\textbf{Planar}} \\
         & $\varepsilon_t$ [$\si{\centi\metre}$] & $\varepsilon_r$ [$\si{\degree}$] & {Time [$\si{\milli\second}$]}& $\varepsilon_t$ [$\si{\centi\metre}$] & $\varepsilon_r$ [$\si{\degree}$] & {Time [$\si{\milli\second}$]}\\
    \midrule
    \makegray{Analytic~\cite{daniilidis1999hand}} & {\makegray{\raisebox{0.2em}{\tiny\textgreater}$10^5$}} & {\makegray{71.5}} & \makegray{4.2} & {\makegray{---}} & {\makegray{---}} & {\makegray{---}}\\
     Matrix~\cite{giamou2019certifiably} & \B 17.84 & 0.263 & 62.7 & {---} & {---} & {---} \\
     Global (ours) & 20.76 & \B 0.257 & 45.1 & \B 15.85 & 0.355 & 53.1\\
     Fast (ours) & 20.76 & \B 0.257 & \B 5.6 & \B 15.84 & \B 0.336 & \B 3.7\\
    \bottomrule    
\end{tabular}
\end{table}

\subsection{Online Estimation}

\begin{figure}
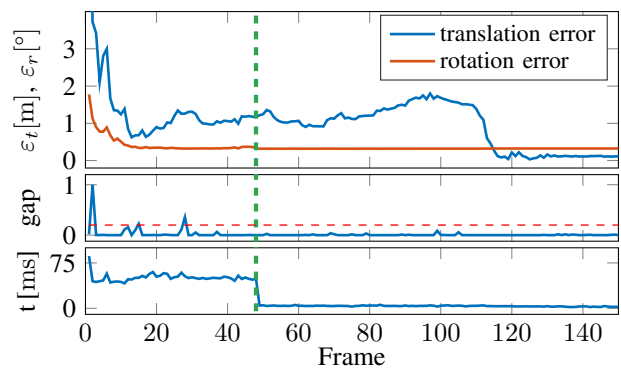

    \centering
    \include{img/online_calib_errors}
    \caption{Rotation and translation error, scaled duality gap, and execution time during an online calibration of sequence 07 of~\cite{geiger2012autonomous} are plotted.
    The red dashed line marks the verification threshold. 
    The green dashed line at 48 frames marks the point from which only the fast approach is invoked, which decreases the execution time significantly.}
    \label{fig:onlineResults}
\end{figure}

For evaluating our online calibration approach, we used sequence~07 of the KITTI dataset~\cite{geiger2012autonomous} and iteratively added the ego-motion estimates for planar motion as described in \ref{sec:onlineCalib}.
Results for the first 150 frames are displayed in Fig.~\ref{fig:onlineResults}.
It shall be noted that, if given more frames, errors converge to the same values as given in Section~\ref{sec:kittiOdometryEval}.
While the fast approach takes \SI{3.7}{\milli\second} for the planar case during offline calibration, the execution time further decreases to \SI{1.8}{\milli\second} during online calibration.
This is comparable to the analytic solution of~\cite{daniilidis1999hand}, which is not able to incorporate previous solutions.
Average verification execution times of approximately \SI{0.05}{\milli\second} can be neglected compared to the over all processing time.

The duality gap indicates discrepancies between the local and the global solution within the first frames.
Therefore, the global approach is used, leading to longer execution times.
Nevertheless, except for the initial step, our online calibration always complies with the sensor update rate of \SI{10}{\Hz}.
As soon as the local approach retrieves verified global results, the global optimization is no longer processed. 
This point is marked with a green dashed line in Fig.~\ref{fig:onlineResults}.
While the rotation error decreases in an exponential manner, the translation error slightly increases first before it rapidly drops.
The motion path displayed in Fig.~\ref{fig:kittiSequences} shows that, due to a turn of the vehicle, motion changes from almost pure translation to additional rotation at this point.
Thereby, the optimization problem gets conditioned more homogeneously, leading to an immediate decrease of translation errors.
Please note that global optimality does not necessarily imply high accuracy.

\subsection{Globality Verification}

Lastly, we evaluated the globality verification of our local solution.
For this, the global solution of an offline calibration was used and falsified intentionally to different degrees.
Comparing the duality gap with the error's magnitude, we observed a strong relation.
Even at small changes of about \SI{0.1}{\degree} or \SI{0.1}{\metre}, discrepancies are reliably indicated.
For online calibration, the duality gap is only related to the respective error magnitude and independent of the frame number since the costs are normalized using the weights $\eta_i$.

%% file: img/surface.tex
\begin{tikzpicture}
    \draw (0, 0) node[inner sep=0] {\includegraphics[width=0.7\columnwidth]{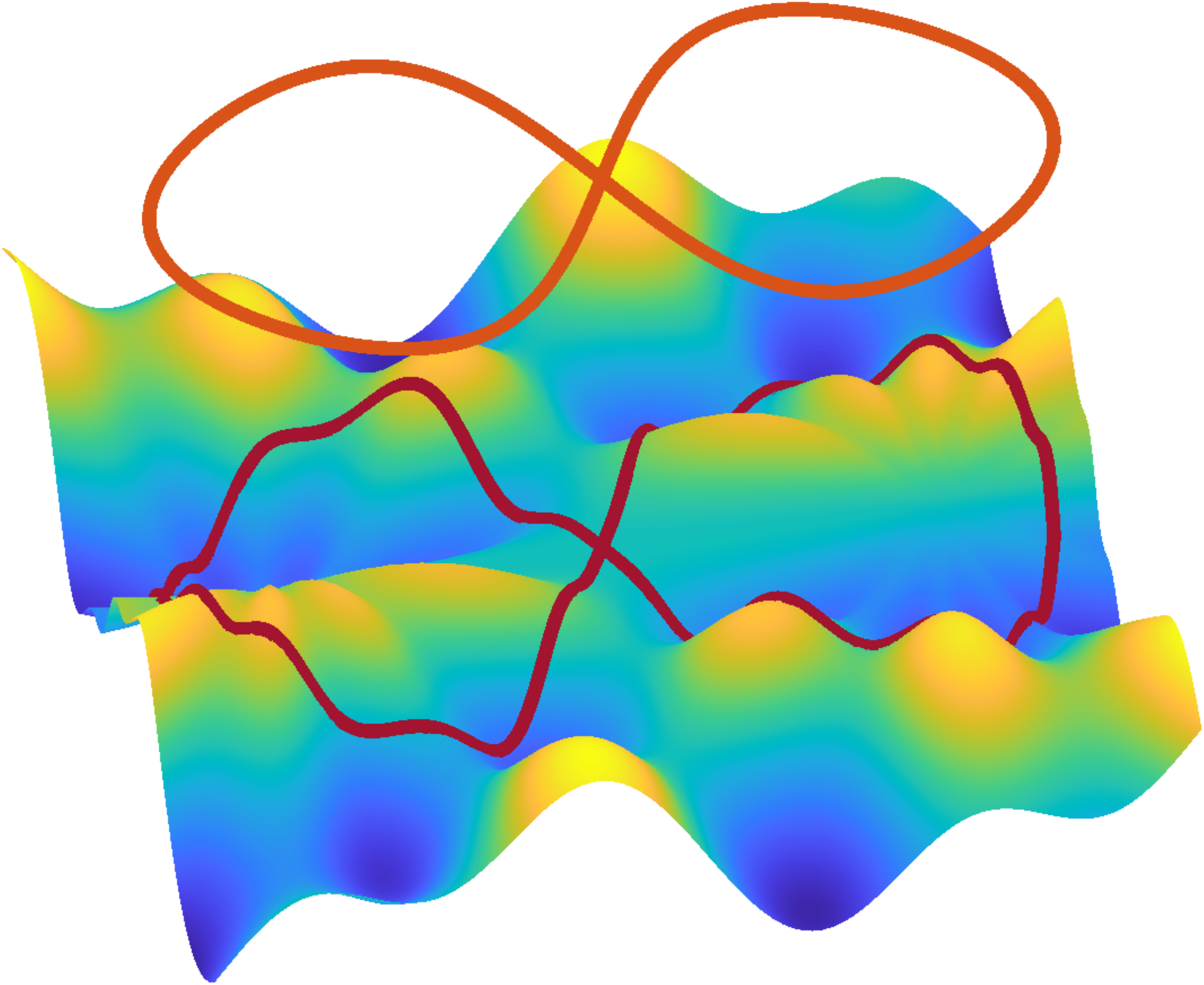}};
    \draw (2.7, 1.8) node {(a)};
    \draw (2.9, 0) node {(b)};
\end{tikzpicture}

%% file: img/translation_calib_errors.tex
\definecolor{mycolor1}{rgb}{0.75294,0.75294,0.75294}%
\definecolor{mycolor2}{rgb}{0.63500,0.07800,0.18400}%
\definecolor{mycolor3}{rgb}{0.00000,0.44700,0.74100}%
\definecolor{mycolor4}{rgb}{0.85000,0.32500,0.09800}%
\definecolor{mycolor5}{rgb}{0.92900,0.69400,0.12500}%
\definecolor{mycolor6}{rgb}{0.49400,0.18400,0.55600}%
\definecolor{mycolor7}{rgb}{0.46600,0.67400,0.18800}%
\begin{tikzpicture}[baseline=(current bounding box.center)]

\begin{axis}[%
width=0.9\linewidth,
height = 0.45\linewidth,
scale only axis,
xmode=log,
xmin=2.5,
xmax=30000,
xminorticks=true,
xlabel style={font=\color{white!15!black}, font=\small},
xlabel={Dataset Size},
ymode=log,
ymin=0.001,
ymax=110,
yminorticks=true,
ylabel style={font=\color{white!15!black}, font=\small,at={(current axis.north west)},above=2mm, rotate=-90},
ylabel={$\varepsilon_r$ [\si{\metre}]},
axis background/.style={fill=white},
title={Simulation Results},
axis x line*=bottom,
axis y line*=left,
xmajorgrids,
xminorgrids,
ymajorgrids,
yminorgrids,
major grid style={black!50},
minor grid style={black!10},
legend columns = 3,
ticklabel style = {font=\tiny},
legend style={font = \small, legend cell align=left, align=left, draw=white!15!black,nodes={scale=0.75, transform shape}}
]

\addplot [color=mycolor2, line width=1.5pt]
  table[row sep=crcr]{%
3	73.5368913115328\\
4	25.8533040180783\\
5	17.351827540399\\
6	15.1078863173534\\
7	14.7917346755809\\
8	11.6047996073766\\
9	10.8130960186226\\
10	10.7184330090139\\
11	10.2312174066764\\
12	8.28785495837127\\
13	6.35458356834902\\
14	8.03518196136197\\
15	7.3488044619873\\
16	7.8997284747883\\
17	5.77473648264647\\
18	6.30705235974431\\
19	6.38722535859981\\
20	8.32789988225344\\
21	6.90677393645842\\
22	6.90045807717266\\
23	5.83839914463688\\
24	5.11097186801145\\
25	5.80667569344623\\
26	5.97609243974579\\
27	5.82305569026625\\
29	5.20356980967487\\
30	4.9864441070679\\
32	7.58094588279203\\
33	5.08103489028105\\
35	5.22350620795786\\
36	5.10947626112411\\
38	3.68480348510009\\
40	3.84656021526539\\
42	3.18377272404931\\
44	3.13332026040346\\
46	2.83714870167137\\
48	4.0449128459879\\
51	3.18140083523093\\
53	4.47321128704943\\
56	3.61789814514478\\
58	3.34147246733586\\
61	3.06553752758066\\
64	3.09528172764429\\
67	2.66385710134707\\
70	2.21008778801211\\
73	2.27201486040493\\
77	2.51054500554372\\
81	2.26113696069317\\
84	2.16305560241057\\
88	2.20873444806403\\
93	2.65965714121887\\
97	2.37481077311466\\
102	2.61303402073471\\
107	2.4451556560584\\
112	1.88089302441231\\
117	2.08301736508097\\
123	2.39138365042287\\
128	2.14995209635752\\
134	1.92455782359855\\
141	1.80132085822967\\
148	2.23005144790942\\
155	2.15609300393609\\
162	2.11756446507395\\
170	1.64919505612074\\
178	1.71705904435727\\
186	1.927085927987\\
195	1.68970425034977\\
204	1.5695106947126\\
214	1.66503976519965\\
224	1.99085482218287\\
235	1.54886886163165\\
246	2.13641257570165\\
257	1.87641386398156\\
270	1.56679304593893\\
282	1.66472806672085\\
296	1.6182336722139\\
310	1.48304816457301\\
325	1.47575544272579\\
340	1.68528450337876\\
356	1.74375065460091\\
373	1.85080411155971\\
391	1.43085319240243\\
409	1.61176097991708\\
429	1.47907066078767\\
449	1.47763976615205\\
470	1.78685743552915\\
493	1.3773593499573\\
516	1.42394858225214\\
540	1.59483450907025\\
566	1.43075772531393\\
593	1.43229998666737\\
621	1.58499788476463\\
651	1.65585772597688\\
681	1.47859053941818\\
714	1.39825713499857\\
747	1.39247945560824\\
783	1.54209689714145\\
820	1.43771426580663\\
859	1.34944488430792\\
900	1.32185498477346\\
942	1.41962644458467\\
987	1.39936923863358\\
1034	1.41465545105604\\
1083	1.40391357591907\\
1134	1.40470737703623\\
1188	1.43972291072191\\
1244	1.41103447258618\\
1303	1.31209396552234\\
1365	1.15121375204472\\
1429	1.34848836844926\\
1497	1.35194742291855\\
1568	1.21632628210881\\
1642	1.36559938875089\\
1720	1.35466113904977\\
1802	1.45553288584563\\
1887	1.30642148032381\\
1976	1.3619425799806\\
2070	1.33980796171439\\
2168	1.43047817413181\\
2271	1.41610714810131\\
2379	1.33662457251545\\
2491	1.34260272709017\\
2609	1.40358229906254\\
2733	1.32979135244057\\
2862	1.34996186634939\\
2998	1.30281802850046\\
3140	1.40967460575717\\
3289	1.36902028891093\\
3445	1.38644246158239\\
3608	1.34955772680234\\
3779	1.34310790354143\\
3958	1.42506465329775\\
4145	1.31752586502357\\
4342	1.33882397748034\\
4547	1.37631831630641\\
4763	1.4248578812191\\
4989	1.30815788450614\\
5225	1.4379202368403\\
5472	1.39208416555109\\
5732	1.40505958193927\\
6003	1.36557989243249\\
6288	1.37683690246608\\
6585	1.36906916777734\\
6897	1.42833624832599\\
7224	1.34779905697692\\
7566	1.3269820567276\\
7925	1.42092775487793\\
8300	1.40354228998745\\
8694	1.3564452926797\\
9105	1.36377017555712\\
9537	1.38882346912372\\
9989	1.35525361623066\\
10462	1.3424868482893\\
10957	1.39216004171256\\
11476	1.35880384707579\\
12020	1.37947586776909\\
12589	1.3593473255654\\
13186	1.37996136614289\\
13811	1.36534071887418\\
14465	1.36312934578788\\
15150	1.34548677141556\\
15868	1.33570166711125\\
16619	1.32850165415276\\
17407	1.38302820288692\\
18231	1.38338142026974\\
19095	1.32984488191929\\
20000	1.35380369615737\\
};
\addlegendentry{100\% noise}

\addplot [color=mycolor3, line width=1.5pt]
  table[row sep=crcr]{%
3	76.1073204018212\\
4	20.8645872503737\\
5	9.39204097121234\\
6	10.2036914338334\\
7	9.10057302403587\\
8	5.23235003110538\\
9	6.58216662927987\\
10	7.83646623983099\\
11	5.76826982067913\\
12	4.87810987781767\\
13	4.27937904019565\\
14	6.63604788720855\\
15	5.23885001745121\\
16	3.46855467796634\\
17	4.25902938090254\\
18	3.35089259176075\\
19	3.39532398785513\\
20	3.89417925125176\\
21	3.72485366720559\\
22	3.18583429733861\\
23	3.11349985476351\\
24	3.09039007619249\\
25	3.59760876091956\\
26	3.37805334338175\\
27	3.07253468546107\\
29	3.33836175822258\\
30	3.37846572977251\\
32	3.34449378173064\\
33	2.93457671101786\\
35	2.38230943038776\\
36	2.1465247009659\\
38	2.40898887545404\\
40	1.75786012114668\\
42	1.82351142949701\\
44	1.88666732597436\\
46	1.57023369794683\\
48	1.8517541213643\\
51	1.63241114612188\\
53	2.97399333464489\\
56	3.91878140246013\\
58	1.84299413963452\\
61	1.47184834209479\\
64	1.49507522323266\\
67	1.01016990584605\\
70	1.31282342474338\\
73	0.959630105592371\\
77	1.22156166754135\\
81	1.27937526826178\\
84	2.24916028191381\\
88	1.0313355667829\\
93	1.31409096076874\\
97	1.30630274406154\\
102	1.09096876255526\\
107	1.08150871519067\\
112	1.2456583335993\\
117	0.918728880769994\\
123	1.02567184398166\\
128	1.80084477207337\\
134	0.866631820355189\\
141	1.07754390989894\\
148	0.897830026873148\\
155	0.847881416192271\\
162	0.954684867437039\\
170	0.988023923071983\\
178	0.861919156998634\\
186	0.740658167874669\\
195	0.892802014970478\\
204	0.783264334630344\\
214	0.675737428770775\\
224	0.93339824819189\\
235	0.853494672997323\\
246	0.874176223530662\\
257	0.779544411244615\\
270	0.728092176947813\\
282	0.692911029783885\\
296	0.706057770410055\\
310	0.755418573087228\\
325	0.560659690842289\\
340	0.564539123873308\\
356	0.576762342382304\\
373	0.558673447219224\\
391	0.517229553591905\\
409	0.72025709584586\\
429	0.544468006736205\\
449	0.589749407886844\\
470	0.518300374989941\\
493	0.557691807440412\\
516	0.687093351861614\\
540	0.577653459517108\\
566	0.547650335519851\\
593	0.761497981370499\\
621	0.584308924662929\\
651	0.481567971544226\\
681	0.623526357696119\\
714	0.50727660706719\\
747	0.518884797181182\\
783	0.604894765578931\\
820	0.749659170437367\\
859	0.424002481415531\\
900	0.518609022278649\\
942	0.513296854445017\\
987	0.533797184163837\\
1034	0.5620232165072\\
1083	0.61192055278162\\
1134	0.410406825312285\\
1188	0.467237628279005\\
1244	0.483462628606402\\
1303	0.502511538529503\\
1365	0.450812989911983\\
1429	0.407609339606583\\
1497	0.418010103069683\\
1568	0.455967014901364\\
1642	0.435474242807696\\
1720	0.489721189715582\\
1802	0.453605062398425\\
1887	0.448772485366207\\
1976	0.469016163571389\\
2070	0.558607564832822\\
2168	0.44180094470288\\
2271	0.417397127924843\\
2379	0.439148444966266\\
2491	0.491336239430709\\
2609	0.407760659694627\\
2733	0.523749620638989\\
2862	0.435113594587361\\
2998	0.43249063862769\\
3140	0.452638832406559\\
3289	0.550618373616338\\
3445	0.481821760118115\\
3608	0.408472748022591\\
3779	0.480566645493075\\
3958	0.42858996037066\\
4145	0.476863428244016\\
4342	0.396709856462389\\
4547	0.398075655044747\\
4763	0.506889399783281\\
4989	0.398392178033694\\
5225	0.489581023337906\\
5472	0.466232435109883\\
5732	0.527777537316299\\
6003	0.473052892138279\\
6288	0.495124204624714\\
6585	0.622512122861747\\
6897	0.521595943534235\\
7224	0.406722704752091\\
7566	0.469102547832947\\
7925	0.55676253190382\\
8300	0.420058703597247\\
8694	0.413447017240972\\
9105	0.400582388219569\\
9537	0.47767646918965\\
9989	0.480885503094607\\
10462	0.397825428377506\\
10957	0.513266244469372\\
11476	0.416076254142215\\
12020	0.473587438575794\\
12589	0.48893894251485\\
13186	0.413482248526784\\
13811	0.554335919788367\\
14465	0.418240559153923\\
15150	0.436660670107868\\
15868	0.458237310239084\\
16619	0.472038081964186\\
17407	0.454687451396648\\
18231	0.393232100309304\\
19095	0.466625137627165\\
20000	0.417360033487505\\
};
\addlegendentry{50\% noise}

\addplot [color=mycolor4, line width=1.5pt]
  table[row sep=crcr]{%
3	98.7519732926123\\
4	10.9466184500392\\
5	3.95241784858348\\
6	3.94161899495193\\
7	3.39630858551508\\
8	2.32605825346327\\
9	2.17755365812775\\
10	2.2003608660991\\
11	2.10091538182797\\
12	2.03581255745824\\
13	1.52928216251149\\
14	1.45173323412897\\
15	2.80748021600634\\
16	1.40201634901581\\
17	1.35322283980218\\
18	1.57810695488298\\
19	1.22663226463587\\
20	1.41307921112022\\
21	1.45912016872779\\
22	1.53827908814156\\
23	1.23455425765726\\
24	1.36009202807583\\
25	1.61909461987709\\
26	1.13413978251675\\
27	1.59090415258569\\
29	1.32658908229043\\
30	1.17583135926253\\
32	0.987017818968996\\
33	1.26773841246498\\
35	1.25197569043863\\
36	0.903398927468434\\
38	1.32071993095718\\
40	0.865809369263756\\
42	0.795097636531849\\
44	0.626011521453354\\
46	0.624780596833453\\
48	0.583479644736886\\
51	0.620823368173592\\
53	0.502684114630821\\
56	0.608743866451901\\
58	0.557541218499572\\
61	0.729468914609335\\
64	0.506455239958085\\
67	0.432052518008654\\
70	0.374317136726054\\
73	0.609942727262672\\
77	0.536576241289475\\
81	0.59039629340766\\
84	0.416829311274661\\
88	0.533310332431915\\
93	0.469565045190654\\
97	0.524724957421757\\
102	0.513393098446656\\
107	0.479593295399603\\
112	0.443145388029078\\
117	0.369891090462658\\
123	0.405458966580376\\
128	0.387435948462684\\
134	0.359931673511602\\
141	0.440067753774118\\
148	0.333535793713395\\
155	0.387378543553408\\
162	0.371392609733525\\
170	0.31541468897905\\
178	0.399317585174117\\
186	0.230471205519983\\
195	0.333358374610658\\
204	0.25506465198134\\
214	0.236326511753587\\
224	0.260093200623185\\
235	0.249971858029378\\
246	0.248744427979132\\
257	0.253683707217985\\
270	0.282538359379528\\
282	0.243962456931776\\
296	0.244923472463401\\
310	0.227839345060792\\
325	0.207476870321527\\
340	0.23398124790122\\
356	0.213355883025632\\
373	0.227519479744961\\
391	0.219758709393843\\
409	0.173421440518088\\
429	0.214458223188622\\
449	0.244153998497474\\
470	0.225826025752132\\
493	0.213968986007599\\
516	0.23516720509863\\
540	0.152070613217367\\
566	0.16968798781561\\
593	0.184612080501002\\
621	0.16533819764294\\
651	0.167556028394772\\
681	0.127416128238268\\
714	0.15686198216086\\
747	0.16486898019623\\
783	0.151786942464486\\
820	0.139821666721104\\
859	0.135566534738846\\
900	0.156288814794943\\
942	0.182607908010432\\
987	0.13449008177433\\
1034	0.131593978612912\\
1083	0.133520814391925\\
1134	0.146613602638591\\
1188	0.120154493628977\\
1244	0.152685624011739\\
1303	0.132571082325295\\
1365	0.126058572614083\\
1429	0.113635606751479\\
1497	0.147907873875793\\
1568	0.112517219637547\\
1642	0.102191184579226\\
1720	0.126952590014941\\
1802	0.142525470150142\\
1887	0.0975261632378255\\
1976	0.110445807158492\\
2070	0.0956838448977445\\
2168	0.110082723879913\\
2271	0.108530948788203\\
2379	0.115007634526288\\
2491	0.082221876549856\\
2609	0.102538416901155\\
2733	0.10097205358982\\
2862	0.104183868521643\\
2998	0.0910315268055841\\
3140	0.10545369455851\\
3289	0.103126832543695\\
3445	0.0855288767077445\\
3608	0.108528252109543\\
3779	0.105546839790023\\
3958	0.0881776028829078\\
4145	0.0929319785343418\\
4342	0.0903729406650015\\
4547	0.106964027370614\\
4763	0.0998091838183236\\
4989	0.0873953614705161\\
5225	0.0798064494197363\\
5472	0.0818479997035054\\
5732	0.0906388999238421\\
6003	0.0863270610605105\\
6288	0.0769012528901854\\
6585	0.0928576391698703\\
6897	0.0793835325720379\\
7224	0.0702357822925366\\
7566	0.0814641018230465\\
7925	0.0871906836595332\\
8300	0.085677618291501\\
8694	0.0816235587866895\\
9105	0.0865905746349064\\
9537	0.0798152428024154\\
9989	0.0717221973348014\\
10462	0.0814764455714887\\
10957	0.0729904854187124\\
11476	0.0768437788705615\\
12020	0.0725490892186583\\
12589	0.076712789007025\\
13186	0.0737889373921645\\
13811	0.0742573854865733\\
14465	0.0766125620804588\\
15150	0.0725510596792849\\
15868	0.0827587817277589\\
16619	0.0799088793538624\\
17407	0.165305189007101\\
18231	0.0809984165995994\\
19095	0.0705256277159577\\
20000	0.0844756438892664\\
};
\addlegendentry{20\% noise}

\addplot [color=mycolor5, line width=1.5pt]
  table[row sep=crcr]{%
3	73.1947663653302\\
4	7.33830358960954\\
5	2.30637263652532\\
6	1.914309439648\\
7	1.4561640231025\\
8	0.885699582093008\\
9	0.947875371933109\\
10	1.07267617370657\\
11	0.982891829298675\\
12	0.89958348742666\\
13	0.812955257738778\\
14	0.786109609640472\\
15	0.655845601737562\\
16	0.761831083091872\\
17	0.834368042439679\\
18	0.601752937458782\\
19	0.767241152497239\\
20	0.816932481101673\\
21	0.695625113850271\\
22	0.6977768586877\\
23	0.783644060582809\\
24	0.769136357658512\\
25	0.791973214437026\\
26	0.784541360647264\\
27	0.573680281409265\\
29	0.77695129532112\\
30	0.560724493715548\\
32	0.627124492919107\\
33	0.488682967785299\\
35	0.55528782032999\\
36	0.467365317098885\\
38	0.442469842738688\\
40	0.449002076755276\\
42	0.371575439379351\\
44	0.360620626038286\\
46	0.324280115593755\\
48	0.251256597918946\\
51	0.292190510067254\\
53	0.325075509168349\\
56	0.303835954171509\\
58	0.318962955282502\\
61	0.310034233224422\\
64	0.226485718085484\\
67	0.284470468208155\\
70	0.257345088260371\\
73	0.239008062924237\\
77	0.274122493518957\\
81	0.231633205274386\\
84	0.275032203552194\\
88	0.187098886471096\\
93	0.264911019429976\\
97	0.179232888478686\\
102	0.212203364157755\\
107	0.226629740850287\\
112	0.237573276177982\\
117	0.263576883302989\\
123	0.199146655596239\\
128	0.168736447736965\\
134	0.174907043468861\\
141	0.191187302233215\\
148	0.143092942139894\\
155	0.140257581042003\\
162	0.125854279520276\\
170	0.196542305056491\\
178	0.161123553476198\\
186	0.173565016016535\\
195	0.143479556293997\\
204	0.137377104855434\\
214	0.114275803778836\\
224	0.10867198391388\\
235	0.146739865197306\\
246	0.144877073577866\\
257	0.102743274324028\\
270	0.0838469829648494\\
282	0.109279833477848\\
296	0.0967563494861147\\
310	0.124755111028067\\
325	0.118767533377235\\
340	0.0889086067186032\\
356	0.0805884009414284\\
373	0.104154157686204\\
391	0.090849633103047\\
409	0.107464953461545\\
429	0.0816716783556637\\
449	0.0889917506658995\\
470	0.0859481171276009\\
493	0.0975640061905242\\
516	0.0979071596160722\\
540	0.100638520635495\\
566	0.0650389258381474\\
593	0.0771118203649354\\
621	0.0591999871676375\\
651	0.0654990078228522\\
681	0.0820789880587324\\
714	0.0708086339916623\\
747	0.0657310795895147\\
783	0.0692574238589699\\
820	0.087106025428789\\
859	0.0677579387401698\\
900	0.0629267345673756\\
942	0.067960169842413\\
987	0.0686780151113919\\
1034	0.0749833846085686\\
1083	0.0531559527729746\\
1134	0.0664557024250204\\
1188	0.0511559679268147\\
1244	0.0539757143299772\\
1303	0.0543614683765355\\
1365	0.0569011982378007\\
1429	0.0540889826798425\\
1497	0.069302315414945\\
1568	0.041001324799145\\
1642	0.0504787616100258\\
1720	0.0540216973111817\\
1802	0.0503601119449477\\
1887	0.0500264724461681\\
1976	0.0542235120195941\\
2070	0.0452089677736976\\
2168	0.0433129203098933\\
2271	0.0477719109624838\\
2379	0.039928538098993\\
2491	0.0477980561102188\\
2609	0.0371440473709123\\
2733	0.0344453079806664\\
2862	0.0417469789210823\\
2998	0.0445665000617964\\
3140	0.036460416139021\\
3289	0.0447652138560473\\
3445	0.0380703491200295\\
3608	0.0376230009753241\\
3779	0.0288169939783558\\
3958	0.0416640444977016\\
4145	0.0385213240614821\\
4342	0.0364144292888499\\
4547	0.0337679360469056\\
4763	0.0279852726628268\\
4989	0.0270825989029448\\
5225	0.032912829320032\\
5472	0.0278020241012649\\
5732	0.0346659536600225\\
6003	0.035991102907052\\
6288	0.0252285644994541\\
6585	0.0278445381253107\\
6897	0.0301911155812297\\
7224	0.0293247186055222\\
7566	0.0273716137813828\\
7925	0.0311464195711814\\
8300	0.030559787159223\\
8694	0.0298295207686527\\
9105	0.0245063556316671\\
9537	0.0253952825171166\\
9989	0.0251860851568216\\
10462	0.0272915044705016\\
10957	0.0295368832442513\\
11476	0.0241826859759923\\
12020	0.0226532939232719\\
12589	0.022925815440162\\
13186	0.0231641494135261\\
13811	0.0216564364766578\\
14465	0.0246365522420806\\
15150	0.0228664590862226\\
15868	0.0191209655948115\\
16619	0.0201092377084197\\
17407	0.0196259078721222\\
18231	0.0260463960221228\\
19095	0.026815648441136\\
20000	0.0207056120575548\\
};
\addlegendentry{10\% noise}

\addplot [color=mycolor6, line width=1.5pt]
  table[row sep=crcr]{%
3	64.5833439344651\\
4	2.34901465703379\\
5	1.26601819962256\\
6	0.690562517020009\\
7	0.736781584617062\\
8	0.514391882113411\\
9	0.530203485614226\\
10	0.425836950888939\\
11	0.540598720784912\\
12	0.528179753845967\\
13	0.32861701075252\\
14	0.447614003957176\\
15	0.373186810951202\\
16	0.462916394418897\\
17	0.326385320660481\\
18	0.292384087488093\\
19	0.395561173270567\\
20	0.360625938785632\\
21	0.307789759742262\\
22	0.307717135388979\\
23	0.334848427903234\\
24	0.280294208240363\\
25	0.334882206146728\\
26	0.294400021883534\\
27	0.418027044680414\\
29	0.352728575224938\\
30	0.286695317643956\\
32	0.322738585497784\\
33	0.243528178979564\\
35	0.277258223319398\\
36	0.180598096730548\\
38	0.296696660289487\\
40	0.170512678928403\\
42	0.25699508022503\\
44	0.209157158648132\\
46	0.155786929081617\\
48	0.173757690235674\\
51	0.141359224583291\\
53	0.153930860590434\\
56	0.157009275009036\\
58	0.157614352200928\\
61	0.107492944351177\\
64	0.124698426424746\\
67	0.12105136375014\\
70	0.127665456456977\\
73	0.118792068053953\\
77	0.124879464910094\\
81	0.0995268981803692\\
84	0.120028610462069\\
88	0.114946994761972\\
93	0.0984458563440205\\
97	0.122584543788711\\
102	0.109547925662558\\
107	0.129985001499768\\
112	0.0990414650294597\\
117	0.0926612804941063\\
123	0.0883027125730381\\
128	0.0996103092347016\\
134	0.0801358334342421\\
141	0.077296364919072\\
148	0.0935978970151835\\
155	0.0744785507542093\\
162	0.0940026031580809\\
170	0.0783757181498693\\
178	0.0628614453447929\\
186	0.0949248149405575\\
195	0.0582660379309902\\
204	0.0637916605035923\\
214	0.0580758768957994\\
224	0.0716120035615209\\
235	0.0613001426900955\\
246	0.0806553388928791\\
257	0.0706969065684846\\
270	0.0509690526946087\\
282	0.0688651173451252\\
296	0.0635176781276753\\
310	0.0649935869681901\\
325	0.0586208147584969\\
340	0.0576157092434207\\
356	0.0536435412307661\\
373	0.0585059629617525\\
391	0.0604782096984881\\
409	0.0549668534131551\\
429	0.0439394961945978\\
449	0.0453048460138808\\
470	0.0565836665262256\\
493	0.0327715901427367\\
516	0.0266990091501125\\
540	0.0411654940454851\\
566	0.036952935492206\\
593	0.0463411844602919\\
621	0.0355421387503234\\
651	0.0352262954716987\\
681	0.040226322390874\\
714	0.0359479824443529\\
747	0.0436161438483604\\
783	0.0338304296041999\\
820	0.0440703148383077\\
859	0.0292933466372886\\
900	0.0282408607657306\\
942	0.0309350730327895\\
987	0.0295480460466882\\
1034	0.0296805708274471\\
1083	0.0292320558407777\\
1134	0.0311194546539149\\
1188	0.0310721042106538\\
1244	0.0197727603457104\\
1303	0.0255539844633588\\
1365	0.0338353719805982\\
1429	0.0241664071642186\\
1497	0.0187369166296553\\
1568	0.0310682783760637\\
1642	0.0297963784288941\\
1720	0.0241818508828715\\
1802	0.0185883346134235\\
1887	0.0206496794894273\\
1976	0.018400632462604\\
2070	0.019559984360364\\
2168	0.0212750052928153\\
2271	0.0210771367770922\\
2379	0.0168375866972447\\
2491	0.0177813866134732\\
2609	0.0196564858044868\\
2733	0.0156145659053835\\
2862	0.0211167705618864\\
2998	0.0174583007618212\\
3140	0.0167416891215821\\
3289	0.0150805172221093\\
3445	0.0154143502491044\\
3608	0.017535339645974\\
3779	0.0149272598918173\\
3958	0.0197796152414669\\
4145	0.0160566893655476\\
4342	0.0164041940143479\\
4547	0.0130690856777373\\
4763	0.0109613615188385\\
4989	0.0157507147474131\\
5225	0.012240991733126\\
5472	0.0144122468378582\\
5732	0.0127578974314809\\
6003	0.0133311710978585\\
6288	0.010005935506126\\
6585	0.0150901544149509\\
6897	0.00981188754129407\\
7224	0.0150582445220694\\
7566	0.011806428539852\\
7925	0.0117526014908352\\
8300	0.0103105540128294\\
8694	0.0105567086922335\\
9105	0.0105932061429031\\
9537	0.0123932461014019\\
9989	0.0125028587756252\\
10462	0.0100849078534148\\
10957	0.00882792119562729\\
11476	0.0113212246644124\\
12020	0.0102099771770799\\
12589	0.0101499027787723\\
13186	0.00938168812837515\\
13811	0.00830817465065484\\
14465	0.0109123372311807\\
15150	0.00936042392957996\\
15868	0.00926136791774801\\
16619	0.00751422184189653\\
17407	0.00887640542149243\\
18231	0.00884780823113759\\
19095	0.00837733720858756\\
20000	0.0088532394808153\\
};
\addlegendentry{5\% noise}

\addplot [color=mycolor7, line width=1.5pt]
  table[row sep=crcr]{%
3	39.8727108159835\\
4	0.61545424123503\\
5	0.16318263546927\\
6	0.161590443078152\\
7	0.170675322240125\\
8	0.118953638547522\\
9	0.0943847685230653\\
10	0.0940766196868938\\
11	0.100249177324863\\
12	0.104243873517798\\
13	0.0734663925229685\\
14	0.0971827572777799\\
15	0.088689221854606\\
16	0.0702325898374892\\
17	0.0845157795270178\\
18	0.0687053755643913\\
19	0.068138254239256\\
20	0.0654515959065956\\
21	0.060263369008008\\
22	0.0644153861458094\\
23	0.0486168463831087\\
24	0.0685913315833316\\
25	0.0514752732229406\\
26	0.0547558078682227\\
27	0.0718339126122009\\
29	0.0874193450679576\\
30	0.0694742454083648\\
32	0.069698319438031\\
33	0.0586189029816401\\
35	0.0577034034973589\\
36	0.0438030711556703\\
38	0.0486500744921114\\
40	0.0439115535731933\\
42	0.0342849653508402\\
44	0.0385704813526558\\
46	0.0360070348757581\\
48	0.0300157421174054\\
51	0.0278950708672095\\
53	0.0266721017125782\\
56	0.0246194752655695\\
58	0.0236952531194513\\
61	0.0316269840033041\\
64	0.0243315187959033\\
67	0.0280375136597436\\
70	0.0289107845610769\\
73	0.029387889012622\\
77	0.0259448444681895\\
81	0.0265117191374944\\
84	0.0238052397807383\\
88	0.021896827980165\\
93	0.0257890527369904\\
97	0.0242971724648493\\
102	0.0305230688228073\\
107	0.031670450570716\\
112	0.0248910957929039\\
117	0.0229069465846994\\
123	0.0200836071001514\\
128	0.0168613794560348\\
134	0.0176446412328544\\
141	0.0188526391799946\\
148	0.013731766512704\\
155	0.0132950384263558\\
162	0.013168552722973\\
170	0.0145844267229154\\
178	0.0123591287490675\\
186	0.0179764893556915\\
195	0.0162862087200623\\
204	0.0132467272344676\\
214	0.0128078873372606\\
224	0.0116909054003307\\
235	0.0140321058944422\\
246	0.0164021341050486\\
257	0.0137238531097932\\
270	0.0108061265704566\\
282	0.0119989976758864\\
296	0.00934402619098975\\
310	0.00930496461688479\\
325	0.0111921376888203\\
340	0.0112819816599495\\
356	0.0114016025327257\\
373	0.00879357902534999\\
391	0.00973703050409026\\
409	0.00797846371355927\\
429	0.00911028140575413\\
449	0.0093021164791922\\
470	0.00892762391859578\\
493	0.0108647588797747\\
516	0.00949098569132118\\
540	0.00878140784959399\\
566	0.00951590201441665\\
593	0.0108331214632109\\
621	0.00795507965388189\\
651	0.00729751807124786\\
681	0.00837316867487549\\
714	0.00595414797936635\\
747	0.00792962133354523\\
783	0.00798689362098996\\
820	0.00724655049785003\\
859	0.00708580436015395\\
900	0.00559871815311264\\
942	0.00759323431454209\\
987	0.00591530927364848\\
1034	0.00761875834637722\\
1083	0.00638456454358501\\
1134	0.00451586960204929\\
1188	0.00582568919841227\\
1244	0.00513038765120764\\
1303	0.00534242607565255\\
1365	0.00602658206440476\\
1429	0.00395267538308521\\
1497	0.00414624883808481\\
1568	0.00513501345465016\\
1642	0.0066780823598839\\
1720	0.0051671194459786\\
1802	0.00491586958225352\\
1887	0.00414147132955146\\
1976	0.00430158383458732\\
2070	0.00401149747983828\\
2168	0.00445714374711273\\
2271	0.00505737869225162\\
2379	0.00412318127279581\\
2491	0.00427497148639197\\
2609	0.00421633970320871\\
2733	0.00408183243292482\\
2862	0.00353460252994266\\
2998	0.00313207380074192\\
3140	0.00358515233997467\\
3289	0.00306082983869465\\
3445	0.00288874435966879\\
3608	0.0035194507827224\\
3779	0.00335616574756716\\
3958	0.00269647481593802\\
4145	0.00282622445721977\\
4342	0.00283898315194406\\
4547	0.00310683606740979\\
4763	0.00295233936263238\\
4989	0.00249792187560101\\
5225	0.0024424955125332\\
5472	0.0027602866829287\\
5732	0.00238807758730266\\
6003	0.00240350094358477\\
6288	0.00257698242092603\\
6585	0.00235372205708436\\
6897	0.00232281738320993\\
7224	0.00252762745533228\\
7566	0.00236599893283885\\
7925	0.00220515287360307\\
8300	0.00233195052601417\\
8694	0.00204240281800918\\
9105	0.00217161398115049\\
9537	0.00184443180881528\\
9989	0.00204661438726748\\
10462	0.00200070368916428\\
10957	0.00178225471300502\\
11476	0.00233531114594898\\
12020	0.00139809696753271\\
12589	0.00186494360250024\\
13186	0.00135708589570067\\
13811	0.00190438197438461\\
14465	0.00135123947220018\\
15150	0.00117352429173882\\
15868	0.00189288335660532\\
16619	0.00152016254546358\\
17407	0.00186585915301768\\
18231	0.00114518808731666\\
19095	0.00154317535427457\\
20000	0.00149535322489713\\
};
\addlegendentry{1\% noise}

\end{axis}
\end{tikzpicture}

%% file: img/kitti_path_07.tex
\definecolor{mycolor1}{rgb}{0.00000,0.44700,0.74100}%
\definecolor{mycolor2}{rgb}{0.46600,0.67400,0.18800}%
\definecolor{mycolor3}{rgb}{0.85000,0.32500,0.09800}%
\begin{tikzpicture}[baseline]

\begin{axis}[%
        width=0.57\linewidth,
        height = 0.23\linewidth,
        scale only axis,
        xmin=-200,
        xmax=215,
        xlabel style={font=\color{white!15!black},yshift=1.5mm},
        xlabel={z [\si{\metre}]},
        ymin=-15,
        ymax=200,
        ylabel style={font=\color{white!15!black}},
        ylabel={-x [\si{\metre}]},
        xmajorgrids,
        xminorgrids,
        ymajorgrids,
        yminorgrids,
        minor tick num = 4,
        major tick style={color=black!50},
        minor tick style={color=black!10},
        major grid style={black!50},
        minor grid style={black!10},
        axis background/.style={fill=white},
        miter limit=1
]
\input{img/kitti_path_07_data}

\addplot [color=mycolor2, only marks, mark=*, mark options={solid, fill=mycolor2, mycolor2}, forget plot]
  table[row sep=crcr]{%
2.220446e-16	-5.551115e-17\\
};
\addplot [color=mycolor3, only marks, mark=*, mark options={solid, fill=mycolor3, mycolor3}, forget plot]
  table[row sep=crcr]{%
8.874404	75.83698\\
};
\end{axis}

\end{tikzpicture}%

%% file: img/online_calib_errors.tex
\definecolor{mycolor1}{rgb}{0.00000,0.44700,0.74100}%
\definecolor{mycolor2}{rgb}{0.85000,0.32500,0.09800}%
\definecolor{mycolor3}{rgb}{0.18039,0.64313,0.30980}%
\definecolor{mycolor4}{rgb}{1,0,0}%
\begin{tikzpicture}[baseline=(current bounding box.center)]

    \begin{groupplot}[
        group style={
            % set how the plots should be organized
            group size=1 by 3,
            % only show ticklabels and axis labels on the bottom
            x descriptions at=edge bottom,
            % set the `vertical sep' to zero
            vertical sep=2.5pt,
            xlabels at=edge bottom
        },
        scale only axis,
        legend cell align={left},
        xlabel style={yshift=1.5mm},
        xlabel={Frame}
    ]
    
    \nextgroupplot[
        width=0.8\linewidth,
        height = 0.235\linewidth,
        scale only axis,
        xmin=0,
        xmax=150,
        ymin=-0.2,
        ymax=4,
        ytick={0, 1,2,3},
        yticklabels={{0},{1}, {2},{3}},
        ylabel style={yshift=0.6mm},
        ylabel={$\varepsilon_t [\si{\metre}]$, $\varepsilon_r [\si{\degree}]$},
        axis background/.style={fill=white},
        legend style={font = \large, legend cell align=left, align=left, draw=white!15!black,nodes={scale=0.75, transform shape}}
    ]
    \input{img/online_calib_plot_data/translation_error_data}
    \addlegendentry{translation error}
    
    \input{img/online_calib_plot_data/rotation_error_data}
    \addlegendentry{rotation error}
    
    \addplot +[mark=none, color=mycolor3, dashed, line width=1.6pt] coordinates {(48, -1) (48, 5)};

    \nextgroupplot[%
        width=0.8\linewidth,
        height = 0.1\linewidth,
        scale only axis,
        scaled y ticks = false,
        xmin=0,
        xmax=150,
        ymin=-3e-09,
        ymax=3e-08,
        ytick={0, 2.5e-08},
        yticklabels={{0},{1}},
        ylabel style={yshift=0.6mm},
        ylabel={gap},
        axis background/.style={fill=white}
    ]
    \input{img/online_calib_plot_data/verification_data}
    \addplot +[mark=none, color=mycolor3, dashed, line width=1.6pt] coordinates {(48, -4e-09) (48, 4e-08)};
    
    \addplot +[mark=none, color=mycolor4, dashed, line width=0.5pt] coordinates {(1, 5e-9) (150, 5e-9)};
    
    \nextgroupplot[
        width=0.8\linewidth,
        height = 0.1\linewidth,
        scale only axis,
        xmin=0,
        xmax=150,
        ymin=-0.01,
        ymax=0.1,
        ytick={0,0.075},
        yticklabels={{0},{75}},
        ylabel style={yshift=-1mm},
        ylabel={t\,[\si{\milli\second}]},
        axis background/.style={fill=white}
    ]
    \input{img/online_calib_plot_data/time_data}
    \addplot +[mark=none, color=mycolor3, dashed, line width=1.6pt] coordinates {(48, -1) (48, 1)};

    \end{groupplot}
    
\end{tikzpicture}%

%% file: doc/6_conclusion.tex
% !TEX root = ../root.tex

\section{Conclusion}

To conclude, we have described two optimization approaches for extrinsic calibration based on per-sensor ego-motion.
Both, the fast local and the global approach are combined for an online calibration estimation with certifiable global solutions.
Our formulation allows to add additional constraints, e.g., for planar motion.
The evaluation on simulated and real-world data demonstrates that our approach achieves state-of-the-art accuracy while simultaneously obtaining the lowest run time on typical data from autonomous vehicles.

For future work, we are planning on adding the scaling factor for monocular visual odometry to our estimation, similar to \cite{zuniga2019automatic, wise2020certifiably}.
Furthermore, we intend to extend our online approach to track sensor calibration drifts and to determine the estimation consistency over time.

%% file: doc/7_appendix.tex
\section*{Acknowledgment}

This research was accomplished within the projects UNICARagil (FKZ\,16EMO0290) and U-Shift\,II (AK\,3-433.62 -DLR/60).
We acknowledge the funding by the Federal Ministry of Education and Research (BMBF) and the State Ministry of Economic Affairs Baden-Württemberg, respectively.